\documentclass[conference]{IEEEtran}
\IEEEoverridecommandlockouts
\usepackage{cite}
\usepackage{amsmath,amssymb,amsfonts}
\usepackage{algorithmic}
\usepackage{graphicx}
\usepackage{textcomp}
\usepackage{xcolor}
\usepackage{booktabs}
\usepackage{makecell}
\usepackage{array}
\usepackage{multirow}
\usepackage{amssymb}
\usepackage{hyperref}
\usepackage{balance}
\usepackage{boldline}
\usepackage{caption}
\usepackage{subcaption}
\usepackage{comment}
\usepackage[most]{tcolorbox}
\usepackage{soul}
\hypersetup{hidelinks}

\usepackage{lipsum}

\newcommand\blfootnote[1]{%
  \begingroup
  \renewcommand\thefootnote{}\footnote{#1}%
  \addtocounter{footnote}{-1}%
  \endgroup
}

\begin{document}

\newcommand{\jz}[1]{{\color{red}{\bf{[JZ:]}} #1}}
\newcommand*{\red}{\textcolor{red}}
\newcommand*{\blue}{\textcolor{blue}}
\newcommand*{\orange}{\textcolor{orange}}
\newcommand*{\green}{\textcolor{green}}

\newcommand{\model}{HAFusion}
\newcommand{\moduleA}{IntraAFL}
\newcommand{\moduleB}{InterAFL}
\newcommand{\moduleC}{ViewFusion}
\newcommand{\moduleD}{RegionFusion}
\newcommand{\fusion}{DAFusion}
\newcommand{\learning}{HALearning}

\DeclareRobustCommand{\IEEEauthorrefmark}[1]{\smash{\textsuperscript{\footnotesize #1}}}

\setlength\abovecaptionskip{1.5pt}
\setlength\belowcaptionskip{1.5pt}
\setlength{\floatsep}{1.5pt}
\setlength{\textfloatsep}{1.5pt}
\setlength{\intextsep}{1.5pt}
\setlength{\abovedisplayskip}{1.5mm}
\setlength{\belowdisplayskip}{1.5mm}
\setlength\abovedisplayshortskip{1.5pt}
\setlength\belowdisplayshortskip{1.5pt}

\definecolor{lgray}{RGB}{227, 228, 224}

\def\BibTeX{{\rm B\kern-.05em{\sc i\kern-.025em b}\kern-.08em
    T\kern-.1667em\lower.7ex\hbox{E}\kern-.125emX}}

\title{Urban Region Representation Learning with Attentive Fusion}

\author{
\IEEEauthorblockN{
Fengze Sun\IEEEauthorrefmark{1}, 
Jianzhong Qi\IEEEauthorrefmark{1}$^\dagger$, 
Yanchuan Chang\IEEEauthorrefmark{1}, 
Xiaoliang Fan\IEEEauthorrefmark{2}, 
Shanika Karunasekera\IEEEauthorrefmark{1}, 
Egemen Tanin\IEEEauthorrefmark{1}}

\IEEEauthorblockA{
\IEEEauthorrefmark{1}\textit{The University of Melbourne}, 
\IEEEauthorrefmark{2}\textit{Xiamen University}
}
\textit{
\{fengzes@student.,  
jianzhong.qi@, 
yanchuanc@student., 
karus@, 
etanin@\}unimelb.edu.au,
fanxiaoliang@xmu.edu.cn}
}

\maketitle

\begin{abstract}
An increasing number of related urban data sources have brought forth novel opportunities for learning urban region representations, i.e., embeddings. The embeddings describe latent features of urban regions and enable discovering similar regions for urban planning applications. 
Existing methods learn an embedding for a region using every different type of region feature data, and subsequently fuse all learned embeddings of a region to generate a unified region embedding.  
However, these studies often overlook the significance of the fusion process. The typical fusion methods rely on simple aggregation, such as summation and concatenation, thereby disregarding correlations within the fused region embeddings. 

To address this limitation, we propose a novel model named \model~\footnote[1]{\label{github}Code and datasets released at \href{https://github.com/MiRuacle24/HAFusion}{https://github.com/MiRuacle24/HAFusion}}. Our model is powered by a dual-feature attentive fusion module named \fusion, which fuses embeddings from different region features to learn higher-order correlations between the regions as well as between the different types of region features. \fusion\ is generic -- it can be integrated into existing models to enhance their fusion process. Further, motivated by the effective fusion capability of an attentive module, we propose a hybrid attentive feature learning module named \learning\ to enhance the embedding learning from each individual type of region features. 
Extensive experiments on three real-world datasets demonstrate that our model \model\ outperforms state-of-the-art models across three different prediction tasks. Using our learned region embeddings leads to consistent and up to 31\% improvements in the prediction accuracy. 

\end{abstract}


\section{Introduction}~\label{sec:introduction}
Urban region representation learning has recently gained much attention in the field of urban data management~\cite{icde1, icde2, icde3, zheng2014urban, zhang2017urban, liu2016urban}, which transforms urban regions into vector representations called \emph{embeddings}.
These embeddings yield the correlation of regional functionalities and valuable insights into urban structures and properties. They are valuable in developing innovative solutions to critical urban issues and common problems in daily life~\cite{icde4, cyc, icde5}. For example, if the manager of a well-run restaurant in a particular region is considering expanding to new locations, utilizing region embeddings can assist in identifying the most comparable regions for this new venture. As the volume of urban data sources continues to grow, it is essential to develop effective methods for region representation learning from rich urban data.

\vspace{-1mm}
\blfootnote{$\dagger$ Corresponding author.}
\vspace{-3mm}
\begin{figure}[t]
     \centering
     \includegraphics[width = 0.48 \textwidth]{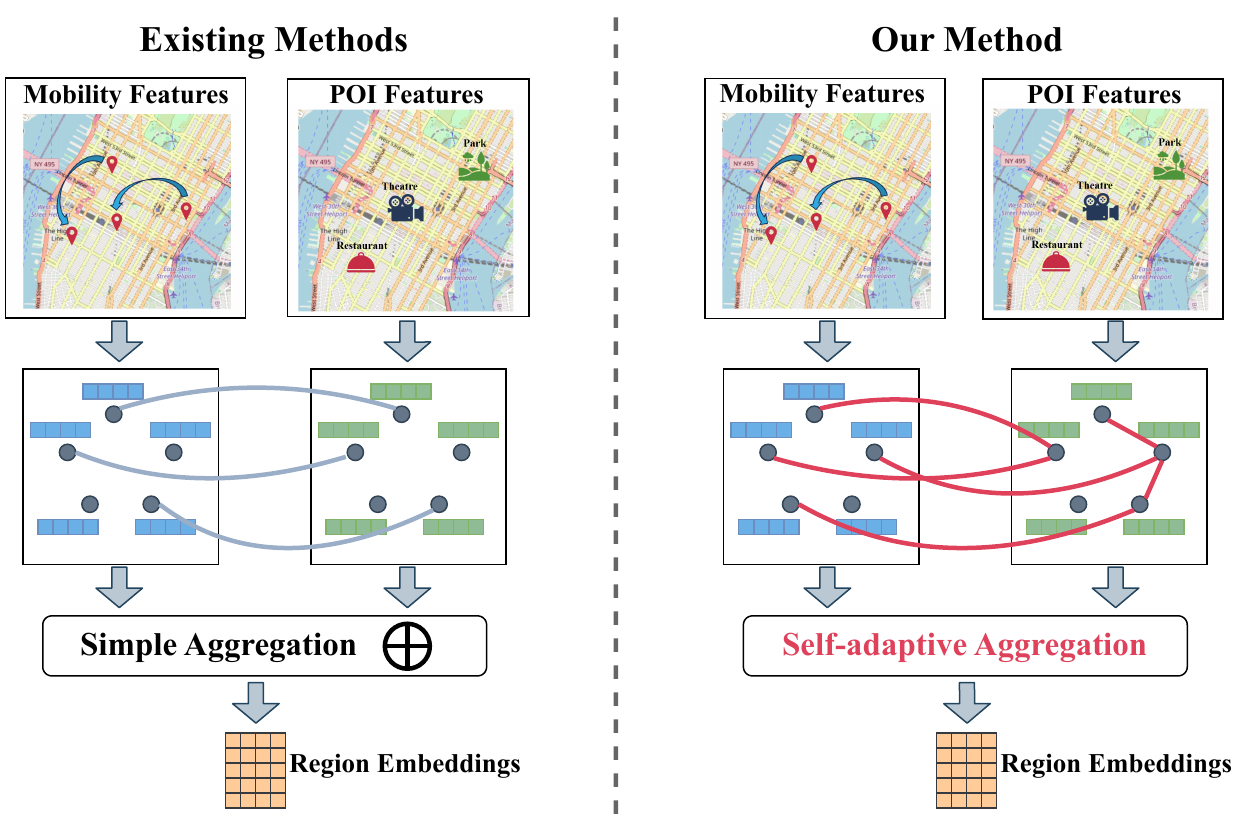}
     \caption{Multi-view fusion region representation learning (each dot denotes a region). Left: existing methods; Right: ours.}
     \label{fig:intro}
\end{figure}

Recently, a rising number of studies \cite{MVURE, MP-VN, CGAL, DLCL, HREP, ReMVC, HUGAT, RegionEncoder, urban2vec, m3g} focus on learning region representations by comparing region correlations from multiple types of region features that describe the same regions (e.g., human mobility and POI features).  Each type of region features is referred to as a \emph{view}, i.e., it depicts the region from a distinct view. For example, two regions may share a similar functionality if they both feature huge human mobility and a large number of entertainment venues.

As Fig.~\ref{fig:intro} (left sub-figure) shows, existing studies~\cite{MVURE, HREP, MP-VN, CGAL} primarily adhere to the following process. Initially, they construct multiple region-based graphs based on distinct views, collectively referred to as \emph{multi-view} graphs. Subsequently, a \emph{view-based embedding} is generated from each individual view graph by capturing \emph{information of the same region} within a single view and across multiple views. Following this, the view-based embeddings are fused to generate a cohesive and unified region embedding.  

There are two limitations with the existing studies.

\textbf{Limitation~1. Existing studies overlook higher-order correlations between regions in the fused embeddings.}
They often emphasize the generation of more effective view-based embeddings from individual views, neglecting the fusion process. The typical fusion methods leverage simple aggregation, such as summation and concatenation, to fuse multiple view-based embeddings. They overlook higher-order correlations between regions in the fused embeddings, while regions are implicitly interconnected through various types of correlations.

\textbf{Limitation~2. Existing studies do not consider the correlations of different regions in different views.}
Existing methods~\cite{MVURE, HREP} typically learn region correlations in the view-based embedding learning process. As Fig.~\ref{fig:intro} (left sub-figure) shows, they consider the correlations of different regions in each individual view, and those of the same region in different views (denoted by the blue curves connecting the same region in different views). They do not consider the correlations of different regions in different views.

To address these limitations, we propose a novel model named~\model~(Fig.~\ref{fig:intro}, right sub-figure), which consists of a \emph{\underline{d}ual-feature \underline{a}ttentive \underline{fusion}} (\fusion) module and a \emph{\underline{h}ybrid \underline{a}ttentive feature \underline{learning}} (\learning)  module. 

\fusion~(Section~\ref{subsec:dualafm}) has two sub-modules: a \underline{view}-aware attentive \underline{fusion} module (\moduleC) and a \underline{region}-aware attentive \underline{fusion} module (\moduleD). \moduleC~aggregates view-based embeddings from multiple views into one, through the attention mechanism. \moduleD~further captures higher-order correlations between regions from the embeddings that have aggregated information from multiple views, through a self-attention mechanism. By introducing \moduleD, we can propagate information in the fused region embeddings to encode more comprehensive region correlation information into the final region embeddings. It is important to note that our \fusion\ module is generic -- it can be integrated with existing models to enhance their fusion process, as shown by our experiments (Section~\ref{sec:exp_dualfusion}). 

\learning~(Section~\ref{sec:hybridafm}) focuses on enhancing the view-based embedding learning process. It captures region correlations in individual views and across views, for \emph{both the same region and different regions} (denoted by the red curves connecting different regions in different views in the right sub-figure of Fig.~\ref{fig:intro}), with \underline{intra}-view \underline{a}ttentive \underline{f}eature \underline{l}earning (\moduleA) and \underline{inter}-view \underline{a}ttentive \underline{f}eature \underline{l}earning  (\moduleB) modules based on adapted self-attention mechanisms. 

To summarize, this paper makes the following contributions:

(1)~We propose an effective learning model named \model~for region representation learning over multiple types of features for the same region.

(2)~For effective learning, we propose two modules:
(i)~The hybrid attentive feature learning module computes view-based embeddings that capture region correlations from both individual views and across different views, for both the same region and different regions.  
(ii)~The dual-feature attentive fusion module fuses the view-based embeddings and further learns higher-order region correlations from the embeddings that have aggregated information from different views. 

(3)~We develop two additional datasets to assess the generalizability of urban region representation learning models.

(4)~We conduct extensive experiments to evaluate \model\ on three real-world datasets. The results demonstrate that \model\ produces embeddings that yield consistent improvements in the accuracy of downstream urban prediction tasks (crime, check-in, and service call predictions), with an advantage of up to 31\% over the state-of-the-art.

\section{Related Work}~\label{sec:relatedwork}
\textbf{Single-view region representation learning.} 
We categorize the single-view region representation learning models by the type of region features considered.

\underline{Human mobility feature-based.} 
\emph{ZE-Mob}~\cite{ZE-Mob} calculates the point-wise mutual information (PMI) between regions by measuring the co-occurrences of regions in the same trip (i.e., as source and destination regions). It learns region embeddings whose dot products approach the PMI values. 
\emph{CDAE}~\cite{CDAE} utilizes human mobility features between POIs to construct a graph in a region. It uses a collective deep auto-encoder model to learn the POI embeddings over the graph, which are aggregated (i.e., weighted sum) as the region embedding.

\emph{MGFN}~\cite{MGFN} constructs multiple mobility graphs, each with trip counts recorded at a different hour. Despite constructing multiple graphs, they are all based on a single type of features. It is thus considered as a single-view approach. After that, it clusters the graphs into seven groups according to their time-weighted ``distances,'' which are defined based on aggregations (e.g., sum) of the edge weights of each graph. All mobility graphs of the same group form a mobility pattern graph. Each mobility pattern graph is then treated as a view, and MGFN learns intra-view and inter-view correlations using vanilla self-attention on these graphs to generate region embeddings. 

\underline{POI feature-based.} 
\emph{HGI}~\cite{HGI} mainly leverage POI features to learn region embeddings. 
It generates embeddings at three hierarchical levels: POI, region (aggregation of POI embeddings), and city (aggregation of region embeddings) levels. It uses a GNN to compute the POI embeddings, which are aggregated to initialize the region embeddings. Another GNN is then used to refine and produce the final region embeddings. 

\underline{Others.} \emph{RegionDCL}~\cite{RegionDCL} uses building footprints. It partitions the buildings in a region into non-overlapping groups using the road network. The footprint of each building (i.e., an image) is used to compute an embedding with a convolutional neural network (CNN).
% powered by contrastive learning~\cite{contrastiveL}. 
The building footprint embeddings of a group are aggregated (average) to initialize the embedding of the group. A contrastive learning layer~\cite{contrastiveL} is applied to refine the building group embeddings. The mean of the embeddings of all building groups in a region is the final region embedding. 
\emph{Tile2Vec}~\cite{tile2vec}, on the other hand,  applies a CNN to learn region embeddings from satellite images directly.

The works above (including our own recent work~\cite{yunxiang}) except for MGFN use single views and do not require multi-view fusion, which differs from our model structure. MGFN constructs multiple views from a single type of input features and thus requires multi-view fusion.  
Our model \model\ also uses self-attention like MGFN does. However, we adapt the self-attention mechanism such that the correlation coefficients learned are explicitly (rather than implicitly as done in MGFN) encoded into the region embeddings learned from each view. Further, we learn the correlation between different regions in different views, instead of just the same region in different views as done in MGFN. Thus, we achieve more effective learning outcomes as shown in Section~\ref{sec:exp}.

\textbf{Multi-view region representation learning} 
Multi-view region representation learning models~\cite{MVURE, MP-VN, CGAL, DLCL, HREP, ReMVC, HUGAT, RegionEncoder, urban2vec, m3g} 
incorporate different types (i.e., views) of region features. 
Studies~\cite{li2018mvstudies, zhao2017mvstudies, zhang2018mvstudies, li2018mvstudies2} have shown that multi-view learning-based models lead to embeddings of a higher quality -- they yield higher accuracy on downstream (prediction) tasks.  
We review these studies based on their view encoders.

\underline{Multi-layer perceptron (MLP)-based.} Methods in this category include \emph{HDGE}~\cite{HDGE}, \emph{MP-VN}~\cite{MP-VN}, \emph{CGAL}~\cite{CGAL} and \emph{ReMVC}~\cite{ReMVC}.
HDGE creates two graphs based on human mobility and the spatial similarity between regions, respectively, where regions are the vertices. The human mobility graph uses the trip count from one region to another (recorded over some period) as an edge weight, and the spatial similarity graph uses the inverse of region distance. HDGE then runs random walks on the two graphs to generate sequences of vertices and applies word2vec~\cite{word2vec} to learn region embeddings.  

MP-VN utilizes spatial distance features and human mobility features (i.e., two views) between POIs to construct two graphs in a region, thereby forming a multi-view solution.
POI embeddings on each graph are randomly initialized and updated with probabilistic propagation-based aggregation~\cite{probability}. The POI embeddings from both graphs are concatenated into a vector to initialize the region embedding. Then, MP-VN uses an MLP-based autoencoder to learn the latent region embedding. 
CGAL further introduces two adversarial modules into the autoencoder, which bring in constraints on the learned embeddings, e.g., the cosine similarities of region embeddings should match the region similarity calculated by POI distributions in the regions. 

ReMVC is based on self-supervised contrastive learning. It uses an MLP-based encoder to transform region features (e.g., count of POIs in different POI categories) into a latent embedding. Then, it performs contrastive learning on embeddings of the same view, as well as embeddings of different views (for the same region). The embeddings of the same region from different views are concatenated as the final region embedding. 

\underline{GNN-based.} GNN-based models~\cite{DLCL,HREP, MVURE, HUGAT} represent regions and their correlations with a graph.
For example, \emph{DLCL}~\cite{DLCL} applies a graph convolutional network (GCN)~\cite{GCN} on top of the region embeddings learned from CGAL, to further learn the correlations between the regions. 

\emph{MVURE}~\cite{MVURE} also treats regions as vertices. It constructs four graphs using human mobility (source and destination), POI category vector similarity, and POI check-in vector similarity for the edges. It then uses graph attention networks (GAT)~\cite{gat} to learn embeddings. The final region embeddings are weighted sums of the embeddings from the graphs.

\emph{HREP}~\cite{HREP} enhances MVURE by replacing the GATs with a relation-aware GCN to learn different relation-specific region embeddings on each graph. Given a downstream task (e.g., crime count prediction), HREP further attaches a task-specific prefix (i.e., a prompt) to the front of each region embedding, which is fine-tuned (i.e., prompt learning) for the task. 
 
\emph{HUGAT}~\cite{HUGAT} constructs a heterogeneous spatio-temporal graph where the vertices include regions, POIs, and time slots. Then, it leverages a heterogeneous GAT~\cite{HAN} to learn region embeddings on the graph.

\underline{Multi-modal-based.} 
Multi-modal-based models~\cite{RegionEncoder,urban2vec,m3g} combine numerical features and visual features (e.g., street view images). 
\emph{RegionEncoder}~\cite{RegionEncoder} uses a GCN and a CNN to jointly learn region embeddings from POIs category count vectors, trip counts, region adjacency data, and satellite images. 
\emph{Urban2Vec}~\cite{urban2vec} and \emph{M3G}~\cite{m3g} use the triplet loss~\cite{triplet} (similar to contrastive learning) to learn region embeddings from street view images and POI textual descriptions.

\textbf{Discussion.} After embeddings are computed from each view, existing models use simple aggregations, e.g., summation or concatenation, to fuse these embeddings and generate the final region embeddings. In contrast, we propose a self-adaptive fusion module that learns higher-order correlations between the regions as well as between the different types of region features, thus obtaining region embeddings that are more effective for downstream (prediction) tasks. 

\begin{figure*}[t]
     \centering
     \vspace{-3mm}
     \includegraphics[width = 0.95 \textwidth]{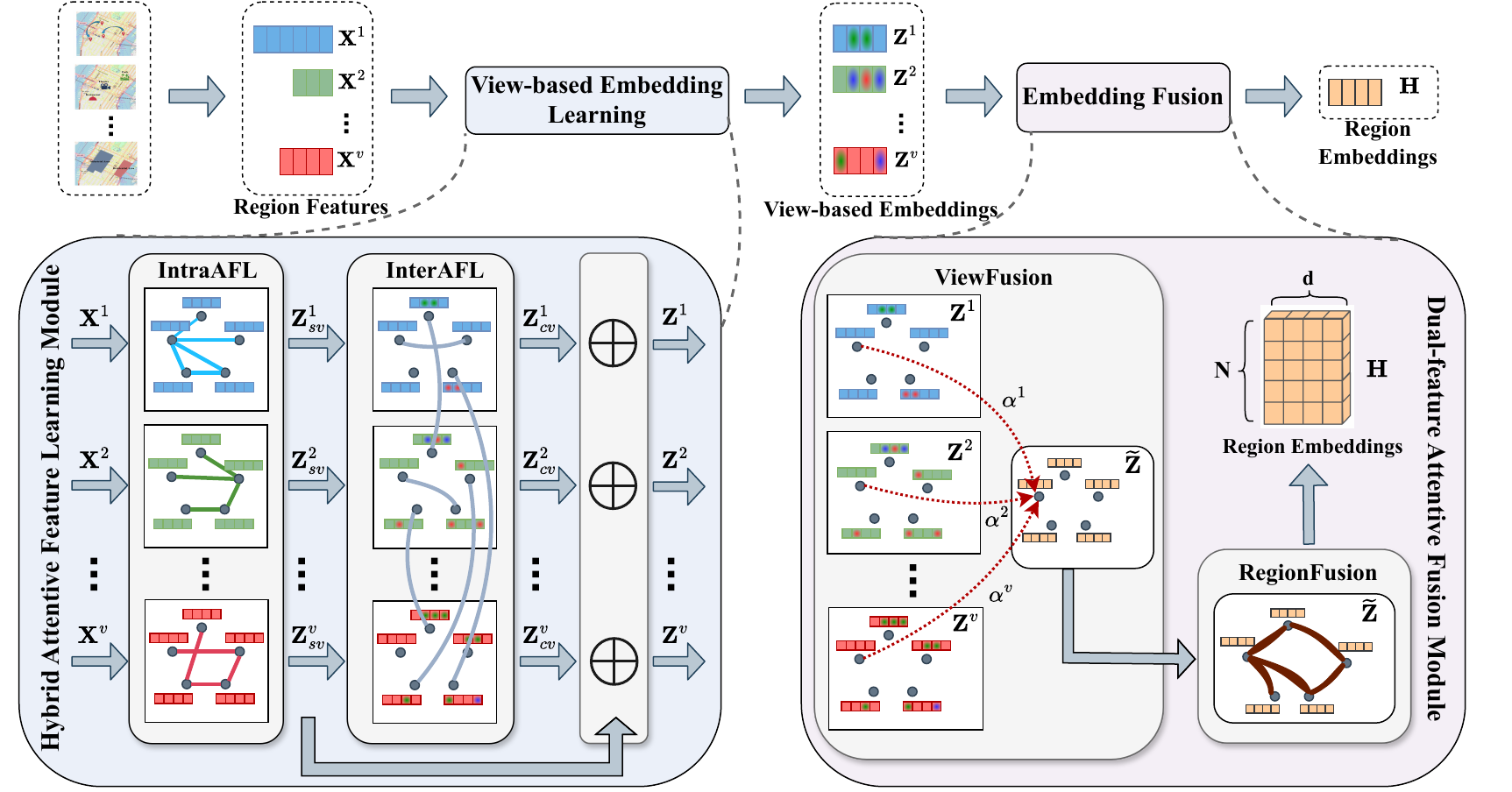}
     \caption{\model\ model overview. 
     The model takes region feature matrices $\mathbf{X}^1, \mathbf{X}^2, 
     \dots, \mathbf{X}^v$  (called views) as input and computes region embeddings with two main modules: (1) The hybrid attentive feature learning module computes view-based embeddings ($\mathbf{Z}^1, \mathbf{Z}^2, \ldots, \mathbf{Z}^v$) for each input view. (2) The dual-feature attentive fusion module further fuses the view-based embeddings at both the view and region perspective to generate the final region embeddings $\mathbf{H}$.}
     \label{fig:modeloverview}
    \vspace{-5mm}
\end{figure*}

\section{Preliminaries}~\label{sec:preliminaries}
We start with a few basic concepts and a problem statement. 
Frequently used symbols are listed in Table~\ref{tab:symbols}.

\begin{table}[ht]
\centering
\renewcommand{\arraystretch}{1.4}
\caption{Frequently Used Symbols}\label{tab:symbols}
\begin{tabular}{c|l}
\hlineB{3}
\textbf{Symbol} & \textbf{Description} \\ \hline \hline
$\mathcal{R}$ & A set of regions (non-overlapping space partitions)  \\ \hline 
$\mathbf{X}^j~(\mathbf{M}, \mathbf{P}, \mathbf{L})$ & Region feature matrices \\ \hline 
$\mathbf{Z}$ & View-based region embedding matrix \\ \hline 
$\widetilde{\mathbf{Z}}$ & Region embedding matrix after ViewFusion \\ \hline 
$\mathbf{H}$ & Final model output region embedding matrix \\ \hline 
\hlineB{3}
\end{tabular}
\end{table}

\textbf{Region.} 
Regions $\mathcal{R}$ refer to a set of non-overlapping space partitions within a designated area of interest, acquired through a certain partition method (e.g., by census tracts). We consider three types of features for each region to compute their representations as follows.

\textbf{Human mobility features.}
Given a set $\mathcal{R}$ of $n$ regions, we define a \emph{human mobility matrix} $\mathbf{M}$ with a set of human flow records between regions, where $m_{i, j} \in \mathbf{M} $ denotes the number of people moving from region $r_i$ to region $r_j$ ($r_i, r_j \in \mathcal{R}$) in an observed period of time  (e.g., a month or a year as in our experimental datasets, which is orthogonal to our problem definition). We further use $\textbf{m}_i$ to denote the vector of outflow records from $r_i$ , i.e., $\textbf{m}_{i} = [m_{i,1}, m_{i,2}, \ldots, m_{i,n}]$. We call it the \emph{human mobility feature vector} of $r_i$.

\textbf{POI features.} 
Given a region $r_i \in \mathcal{R}$, we collect all POIs in $r_i$ from OpenStreetMap~\cite{osm} and extract the POI categories. We consider 26 categories (e.g., restaurants, schools, etc.) following a previous study~\cite{yunxiang} and count the number of POIs in each category, resulting in a 26-dimensional \emph{POI feature vector} $\mathbf{p}_{i}$. Such vectors of all $n$ regions form a \emph{POI matrix} $\mathbf{P}$. 

\textbf{Land use features.} 
The land use features share a similar setup with the POI features. Given a region $r_i$, we count the number of zones in $r_i$ that fall into different zone categories, e.g., industrial zones, residential zones, or transportation zones. These counts together form the \emph{land use feature vector} $\mathbf{l}_i$ of $r_i$. The land use feature vectors of all $n$ regions form a \emph{land use matrix} $\mathbf{L}$.

Human mobility and POI features have been commonly used in urban region representation learning~\cite{MP-VN, CGAL, RegionEncoder, MVURE, HREP}, while we are the first to introduce land use features into the problem. Compared with the POI features, the land use features are coarser-grained, which could yield a clearer overall picture about the most distinctive functionalities of a region. Land use data is often available in the open data repositories of different cities~\cite{nycOpendata, chiOpendata, sfOpendata}.

\textbf{Problem statement.} 
Given a set $\mathcal{R}$ of regions and $v$ $(v > 1)$ feature matrices $\mathbf{X}^1, \mathbf{X}^2, \ldots, \mathbf{X}^v$ of the regions, we aim to learn a function $f$~:~$(r_i, \mathbf{x}^1_{i}, \mathbf{x}^2_{i}, \ldots, \mathbf{x}^v_{i}) \rightarrow \mathbf{h}_{i}$ that maps a region $r_i \in \mathcal{R}$, described by its features $\mathbf{x}_{j} \in \mathbf{X}^j$ $(j \in [1, v])$ , to a $d$-dimensional vector $\mathbf{h}_{i} \in \mathbf{H}$, where $d$ is a small constant. 

The learned region representation (i.e., $\mathbf{h}_{i}$) is expected to preserve characteristics of the region and can be applied to a wide range of downstream tasks, e.g., crime count predictions.

Here, the human mobility, POI, and land use matrices $\mathbf{M}$, $\mathbf{P}$, and $\mathbf{L}$ serve as the three feature matrices  $\mathbf{X}^1$,  $\mathbf{X}^2$, and  $\mathbf{X}^3$ in our model implementation. For generality of discussion, we will use  $\mathbf{X}^j$ to refer to an input region feature matrix.  
\section{Proposed model}
This section presents our proposed model~\model.  
We first provide an overview of the model~(Section~\ref{subsec:modeloveriew}). Then, we detail its core component, the \emph{dual-feature attentive fusion} module (Section~\ref{subsec:dualafm})  
Finally, we present the training objective of our model (Section~\ref{subsec:objective}). 

\vspace{-1mm}
\subsection{Model Overview}~\label{subsec:modeloveriew}
% \vspace{-1mm}
Fig.~\ref{fig:modeloverview} shows the overall architecture of \model.
The model takes as input a set $\mathcal{R}$ of regions as represented by their feature matrices $\mathbf{X}^1, \mathbf{X}^2, \ldots, \mathbf{X}^v$. 
We call each feature matrix a \emph{view} as it corresponds to a different type of features.  
The views are first fed into a feature learning module to learn \emph{view-based region embeddings} on each view. Any feature learning model can be applied, e.g., a GNN as done in the literature~\cite{MVURE, HREP}.
We propose a  \underline{h}ybrid \underline{a}ttentive feature \underline{learning} module (\learning) for the task (Section~\ref{sec:hybridafm}). 

The view-based region embedding matrices, denoted by $\mathbf{Z}^1, \mathbf{Z}^2, \ldots, \mathbf{Z}^v$, are then fed into our \underline{d}ual-feature \underline{a}ttentive \underline{fusion}  (\fusion) module to generate the final region embeddings $\mathbf{H}$.
\fusion\ fuses (1) embeddings of the same region from different views and (2) embeddings of different regions (hence ``dual fusion''). It captures the correlation and dissimilarity between the regions. The generated region embeddings can be used in various downstream tasks, e.g., crime count and check-in predictions as shown in our experimental study. 

It is important to note that our model \model\ offers a generic framework to learn region embeddings with multiple (not necessarily our three) input features. As will be shown in our experiments (Section~\ref{sec:exp_dualfusion}), our \fusion\ module can be easily integrated into existing models~\cite{HREP, MVURE, MGFN} and enhance their fusion process, resulting in region embeddings of higher effectiveness for downstream prediction tasks. 

\vspace{-0.5mm}
\subsection{Dual-Feature Attentive Fusion}~\label{subsec:dualafm}
\begin{figure}[h]
     \centering
     \includegraphics[width = 0.44 \textwidth]{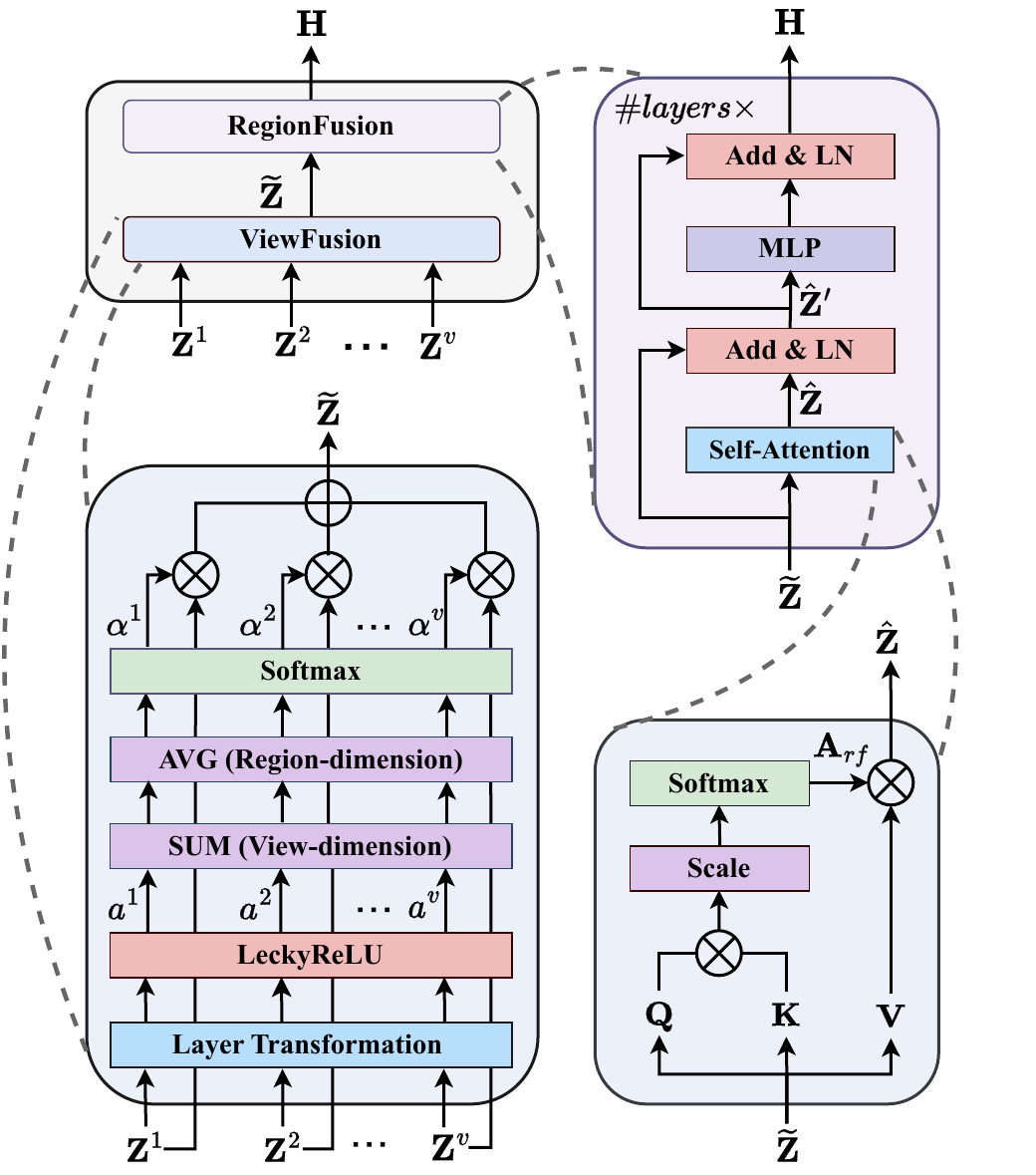}
     \caption{Dual-feature attentive fusion. (1)~\fusion\ (upper left); (2) \moduleC~(lower left); (3)~\moduleD~(right). }
     \label{fig:dualfusion}
     \vspace{-1mm}
\end{figure}

As Fig.~\ref{fig:dualfusion} shows, \fusion\ has two sub-modules: \emph{view-aware attentive fusion}  (\moduleC) and \emph{region-aware attentive fusion} (\moduleD). \moduleC\ fuses the view-based embeddings of the same region from different views into one embedding;  \moduleD\ further fuses the resulting embeddings of different regions to learn their higher-order correlations and derive the final embedding of each region.

\textbf{View-aware attentive fusion.}
\moduleC\ leverages the attention mechanism to learn  \emph{fusion weights} that indicate the relative importance of the views to aggregate the view-based embeddings \{$\mathbf{Z}^1, \mathbf{Z}^2, ..., \mathbf{Z}^v\}$.
It first computes correlation scores between different views by adopting a simplified self-attention~\cite{gat} as follows: 
\vspace{-1mm}
\begin{equation}
a^{jk}_{i} = \mathrm{LeakyReLU}\big(\mathbf{a}^\intercal(\mathbf{W}_F\mathbf{z}^j_{i}||\mathbf{W}_F\mathbf{z}^k_{i})\big),
\end{equation}
where $a^{jk}_{i}$ is the correlation score between the $j$-th and the $k$-th views of the region $r_i$,  $\mathbf{a} \in \mathbb{R}^{2d'}$ and $\mathbf{W}_F \in \mathbb{R}^{d'\times d}$ are learnable parameters (Linear Transformation in Fig.~\ref{fig:dualfusion}), $d'$ is the dimensionality of the latent representation ($d' = 64$ in the experiments),  `$||$' denotes concatenation, and LeakyReLU is an activation function.

Then, we aggregate the correlation scores along the views and the regions to obtain an overall weight for each view, and then we apply a Softmax function to obtain the normalized fusion weight $\alpha^j$ ($j \in [1, v]$) of each view.
\vspace{-2mm}
\begin{equation}
\alpha^j = \mathrm{Softmax}(\frac{1}{n} \sum^{n}_{i = 1} \sum^{v}_{k=1} a_{i}^{jk}),
\end{equation}
\vspace{-2mm}

We use the fusion weights to fuse the view-based embeddings into a single embedding matrix, denoted as $\widetilde{\mathbf{Z}} \in \mathbb{R}^{n\times d}$:
\begin{equation}
\widetilde{\mathbf{Z}} = \sum_{j=1}^{v}\alpha^j \cdot \mathbf{Z}^j
\end{equation}

\textbf{Region-aware attentive fusion.}
\moduleD~further applies self-attention~\cite{attention} on the embeddings $\widetilde{\mathbf{Z}}$ learned by \moduleC, to encode the higher order correlations among the learned region embeddings. 

As Fig.~\ref{fig:dualfusion} shows, the embeddings $\widetilde{\mathbf{Z}}$ are first fed into a self-attention module, which applies linear transformations to  $\widetilde{\mathbf{Z}}$ to form three projected matrices in latent spaces, i.e., a query matrix $\mathbf{Q} \in \mathbb{R}^{n \times d} = \mathbf{W}_Q \widetilde{\mathbf{Z}}$, a key matrix $\mathbf{K} \in \mathbb{R}^{n \times d} = \mathbf{W}_K \widetilde{\mathbf{Z}}$, and a value matrix $\mathbf{V} \in \mathbb{R}^{n \times d} = \mathbf{W}_V \widetilde{\mathbf{Z}}$. Here, $\mathbf{W}_Q$, $\mathbf{W}_K$, and $\mathbf{W}_V$ are learned parameter matrices of the linear transformations and are all in $\mathbb{R}^{d \times d}$. Multi-head  attention~\cite{attention} is applied here to enhance the model learning capacity, while its details are omitted as it is a direct adoption.

Next, we compute attention coefficients between regions: 
\begin{equation}\label{eq:attn_coeffients}
\mathbf{A}_{rf} = \mathrm{Softmax}\big(\frac{\mathbf{Q} \cdot {\mathbf{K}}^\intercal}{\sqrt{d}}\big),
\end{equation}
where $\sqrt{d}$ is a scaling factor and $\mathbf{A}_{rf} \in \mathbf{R}^{n \times n}$ is a coefficient matrix that records the correlation between every two regions.

After that, we compute hidden representations of the regions based on the attention coefficients $\mathbf{A}_{rf}$:
\begin{equation}\label{eq:attn_c}
    \hat{\mathbf{Z}} = \mathbf{A}_{rf} \cdot \mathbf{V}.
\end{equation}

The output of the self-attention module $\hat{\mathbf{Z}}$ is added with the input embeddings $\widetilde{\mathbf{Z}}$ (through a residual connection) and then goes through a layer normalization (LN) with dropout (to alleviate issues of exploding and vanishing gradients). Subsequently, a multi-layer perceptron (MLP) and another layer normalization (again with a residual connection) are applied to enhance the model learning capacity as follows:
\begin{gather}
    \hat{\mathbf{Z}}' = \mathrm{LayerNorm}(\widetilde{\mathbf{Z}} + \mathrm{Dropout}(\hat{\mathbf{Z}})), \label{eq:attn_postprocess1} \\
    {\mathbf{H}} = \mathrm{LayerNorm}(\hat{\mathbf{Z}}' + \mathrm{Dropout}(\mathrm{MLP}(\hat{\mathbf{Z}}'))). \label{eq:attn_postprocess2}
\end{gather}
Here, $\hat{\mathbf{Z}}'$ is the output of the first layer normalization, and $\mathbf{H}$ is the output region embeddings. 

We stack multiple layers of the structure above, and the output of the final layer is our learned region embeddings $\mathbf{H}$.

\vspace{-1mm}
\subsection{Model Training}~\label{subsec:objective}
% \vspace{-2mm}
We present a multi-task learning objective $\mathcal{L}$ with $v$ sub-objective functions $\mathcal{L}^1, \mathcal{L}^2, \ldots, \mathcal{L}^v$ to guide our model \model\ to learn generic and robust region representations: $\mathcal{L} = \mathcal{L}^1 + \mathcal{L}^2 + \ldots + \mathcal{L}^v$.
Here, each sub-objective function $\mathcal{L}^j$ ($j \in [1, v]$) focuses on learning from an input feature.

Given that there are no external supervision signals (e.g., region class labels), we use the similarity between the regions as computed by their input feature vectors to guide model training. To compute $\mathcal{L}^j$, we use an MLP (i.e., a linear layer and a ReLU activation function)  to map the region embedding matrix $(\mathbf{H})$ to an input-feature-oriented region embedding matrix $(\mathbf{H})^j$, i.e., $\mathbf{H}^j = \mathrm{MLP}(\mathbf{H})$. Then, 
\begin{equation}\label{eq:sub_objective}
    \mathcal{L}^j = \frac{1}{n}\frac{1}{n}\sum_{i=1}^{n}\sum_{k=1}^{n}\Big|\mathrm{cos}(\mathbf{x}^j_i, \mathbf{x}^j_k) - {\mathbf{h}^{j}_{i}} \cdot \mathbf{h}^j_{k}\Big|
\end{equation}
Here, $\mathbf{x}^j_i$ and $\mathbf{x}^j_k$ are  feature vectors of regions $r_i$
 and $r_k$ of the $j$-th input feature, and $\mathrm{cos}(\cdot)$  computes their cosine similarity. Vectors $\mathbf{h}^{j}_{i}$ and  $\mathbf{h}^j_{k}$ from $\mathbf{H}^j$ are the learned embeddings of $r_i$ and $r_k$ mapped towards the $j$-th feature, and their dot product represents the region similarity in the embedding space. The intuition is that the learned embeddings should reflect the region similarity as entailed by the input features. 
We note that the cosine similarity of $\mathbf{h}^j_i$ and $\mathbf{h}^j_k$ can be used in Equation~\ref{eq:sub_objective} instead of the dot product. Empirically, we find that both approaches produce embeddings that yield very similar accuracy in downstream tasks, while using the cosine similarity takes more time in embedding learning. Thus, we have used the dot product instead.

The loss function above is generic and can be applied to input features without requiring further domain knowledge. We use it for the POI and land use features. 
For the human mobility features, they entail both the source and the destination distribution patterns of human mobility. While the sub-objective function above also works, we follow previous studies~\cite{MVURE, HREP, MGFN} and use a  \emph{KL-divergence} based sub-objective function described below. 

\textbf{Mobility distribution loss.} Using the human mobility feature matrix $\mathbf{M}$ (cf.~Section~\ref{sec:preliminaries}). We compute two transition probabilities for movements from region $r_i$ to region $r_k$: 

\vspace{-2mm}
\begin{equation}
\begin{aligned}
    p_s(r_k | r_i) = \frac{m_{i, k}}{\sum^n_{l=1} m_{i, l} }  ,
\end{aligned}
\quad
\quad
\begin{aligned}
    p_d(r_k | r_i) = \frac{m_{i, k}}{\sum^n_{l=1} m_{l, k} } 
\end{aligned}
\end{equation}
\vspace{-2mm}

Here, $p_s(\cdot)$ denotes how likely people move to region $r_k$ when they move out from region $r_i$, and $p_d(\cdot)$ denotes how likely people come from  $r_i$ when they move into  $r_k$.

To enable computing the KL-divergence loss, we need such probability values from the learned embeddings. We map $\mathbf{H}$ into a source feature-oriented matrix $\mathbf{H}^S$ and a destination feature-oriented $\mathbf{H}^D$ (each with an MLP like above).  We compute two transition probabilities with these matrices: 
\begin{equation}
\hat{p_s}(r_k \vert r_i) = \frac{\exp(\mathbf{h}^S_{i} \cdot \mathbf{h}^D_{k})}
{\sum^n_{l=1} \exp(\mathbf{h}^S_{i} \cdot \mathbf{h}^D_{l})} 
\end{equation}

\begin{equation}
\hat{p_d}(r_k \vert r_i) = \frac{\exp(\mathbf{h}^S_{i} \cdot \mathbf{h}^D_{k})}
{\sum^n_{l=1} \exp(\mathbf{h}^S_{l} \cdot \mathbf{h}^D_{k})} 
\end{equation}
Here, $\mathbf{h}^S_{i} \in \mathbf{H}^S$ and $\mathbf{h}^D_{k} \in \mathbf{H}^D$ correspond to regions $r_i$ and $r_k$, respectively. The KL divergence loss is then computed:
\begin{equation}
\begin{aligned}
    \mathcal{L}^M  = \sum_{i=1}^{n}\sum_{k=1}^{n} \big(
        & -p_s(r_k \vert r_i)\log(\hat{p_s}(r_k \vert r_i)) \\ 
        & - p_d(r_k \vert r_i)\log(\hat{p_d}(r_k \vert r_i)) 
    \,\big).
\end{aligned}
\end{equation}

\section{Hybrid Attentive Feature Learning}~\label{sec:hybridafm}
Motivated by the strong learning performance of our attention-based fusion module, we further propose an attention-based embedding learning module for the view-based embedding learning process, i.e., the \emph{hybrid attentive feature learning}  (\learning) module, as mentioned in Section~\ref{subsec:modeloveriew}.

\learning\ consists of an \emph{intra-view attentive feature learning} (\moduleA) module and an \emph{inter-view attentive feature learning}  (\moduleB)  module. \moduleA~encodes region correlation entailed in the input features of each single view, while \moduleB~further captures region correlation across views.

\begin{figure}[h]
     \centering
     \includegraphics[width = 0.4 \textwidth]{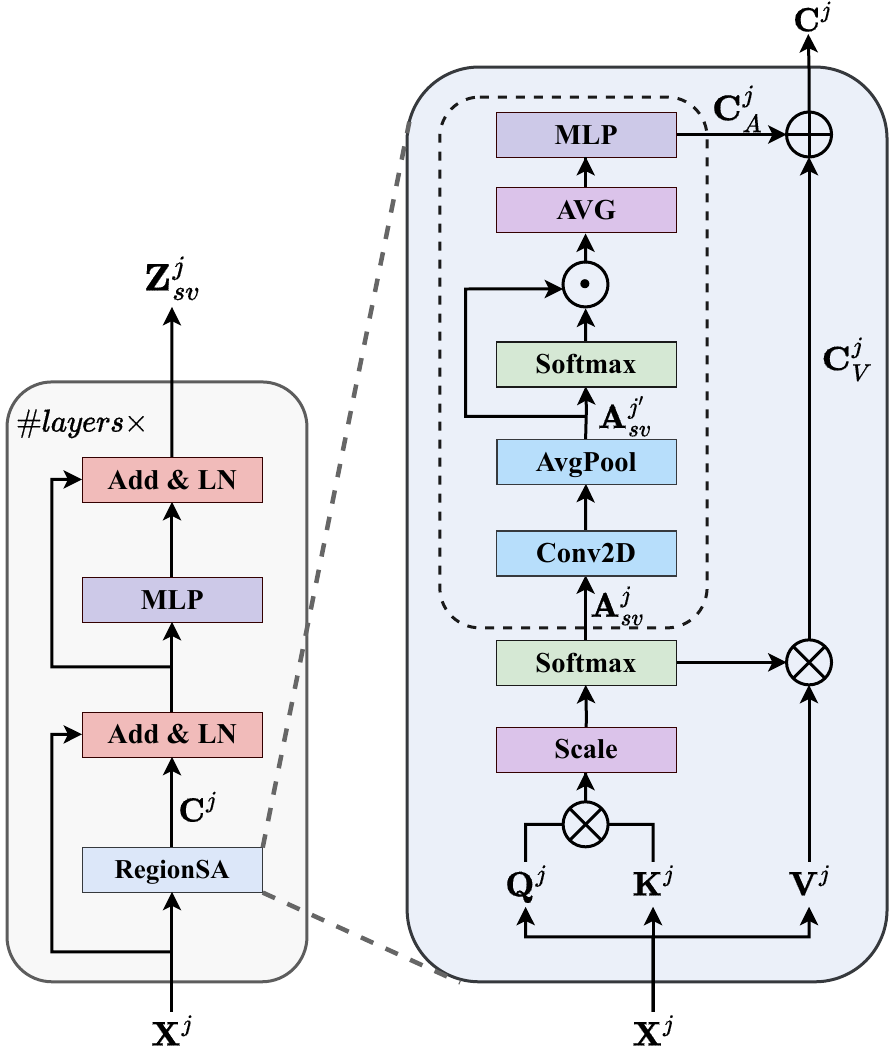}
     \caption{Intra-view attentive feature learning (\moduleA).}
     \label{fig:singleviewafm}
\end{figure}

\textbf{Intra-view attentive feature learning.} Fig.~\ref{fig:singleviewafm} shows the structure of the \moduleA\ module, which is also based on the Transformer encoder~\cite{attention} structure like the \moduleD\  module described earlier, except that its input is a view matrix $\mathbf{X}^j$ and its output is an intermediate region embedding matrix, denoted by  $\mathbf{Z}^j_{sv}$ (``$sv$'' refers to ``single view''). 

We do not repeat the full computation steps of \moduleA. Instead, we focus on its \emph{RegionSA} module (right sub-figure in Fig.~\ref{fig:singleviewafm}), which is our new design for learning both region correlations and region embeddings within each view.

RegionSA adapts the vanilla multi-head self-attention by incorporating a simple yet effective module (inside the dashed rectangle in Fig.~\ref{fig:singleviewafm}). It takes as input the region features of one view,  $\mathbf{X}^j$, and computes the attention coefficient scores between the regions, denoted by  $\mathbf{A}^j_{sv}$, and the latent region representations, denoted by $\mathbf{C}^j_V$.
The calculations are similar to Equations~\ref{eq:attn_coeffients} and~\ref{eq:attn_c} except that the inputs to the equations are different. We hence do not repeat the equations again. 

Next, we aim to combine the intermediate view-based region embeddings $\mathbf{C}^j_V$ and the coefficient score matrix  $\mathbf{A}^j_{sv}$, such that the region embeddings further encode the region correlation information. We observe that it is sub-optimal to directly concatenate the two matrices. This is because the coefficient score matrix $\mathbf{A}^j_{sv}$ is obtained by computing the correlation between every two regions. It does not capture the correlations among multiple regions, e.g., the feature of a region may be impacted by multiple regions jointly. 

To address this issue, we introduce a lightweight module (inside the dashed rectangle in Fig.~\ref{fig:singleviewafm}) applied on $\mathbf{A}^j_{sv}$ to learn multi-region correlations.
We first employ a 2-dimensional convolution layer and an average pooling layer, denoted as Conv2D and AvgPool, to compute a correlation matrix (in $\mathbb{R}^{c \times n \times n}$) that learns the higher-order region correlations (entailed in convolution operation), where $c$ is a hyper-parameter ($c=32$ in our experiments).
Let the resulting matrix be ${\mathbf{A}^{j'}_{sv}}$. 
We compute the element-wise product between 
${\mathbf{A}^{j'}_{sv}}$ and its normalized version (obtained by a Softmax layer). Finally, we average the resulting matrix and project the matrix to $\mathbb{R}^{n \times d}$ with an MLP, such that it can be added with $\mathbf{C}^j_V$. 
The steps above are summarized as follows:
\begin{equation}
{\mathbf{A}^{j'}_{sv}} = \mathrm{AvgPool}(\mathrm{Conv2D}({\mathbf{A}^j_{sv}})),
\end{equation}
\begin{equation}
\mathbf{C}^j_A = \mathrm{MLP}(\mathrm{AVG}({\mathbf{A}^{j'}_{sv}} \odot \mathrm{Softmax}({\mathbf{A}^{j'}_{sv}}))).
\end{equation}

Then, we encode the region correlation information into the region embeddings as the output of RegionSA, denoted as $\mathbf{C}^j$:
\begin{equation}
\mathbf{C}^j = \mathbf{C}^j_V + \mathbf{C}^j_A.
\end{equation}

Matrix $\mathbf{C}^j$ will go through the 
rest of the computation steps of \moduleA\ as shown in the left  sub-figure of Fig.~\ref{fig:singleviewafm} (which resemble Equations~\ref{eq:attn_postprocess1} and \ref{eq:attn_postprocess2}) to produce the output embeddings of the module, 
$\mathbf{Z}^j_{sv}$. Overall, from the $v$ input views, we will obtain $v$ embedding matrices, $\mathbf{Z}_{sv}^1, \mathbf{Z}_{sv}^2, \ldots,  \mathbf{Z}_{sv}^v$.
We call these matrices the \emph{intra-view attentive region embedding matrices}.

\textbf{Inter-view attentive feature learning.}
Next, we present \moduleB\ to learn correlations between regions across views. Note that we consider the correlations between all regions across all views, instead of only the same region from different views as done in existing works~\cite{MVURE, ReMVC, HREP, MGFN}. 

A naive method is to apply self-attention over 
$\mathbf{Z}_{sv}$, like \moduleD\ does, to compute correlation coefficients between all pairs of regions from the same and different views. This approach, however, is computationally expensive, as 
$\mathbf{Z}_{sv}^1, \mathbf{Z}_{sv}^2, \ldots,  \mathbf{Z}_{sv}^v$ together form a large matrix in $\mathbb{R}^{n\times v \times d}$, and it introduces unnecessary noise signals -- many regions far away from each other are unlikely to be correlated.  

\begin{figure}[h]
     \centering
     \includegraphics[width = 0.32 \textwidth, trim=2.5cm 0 0 0 ]{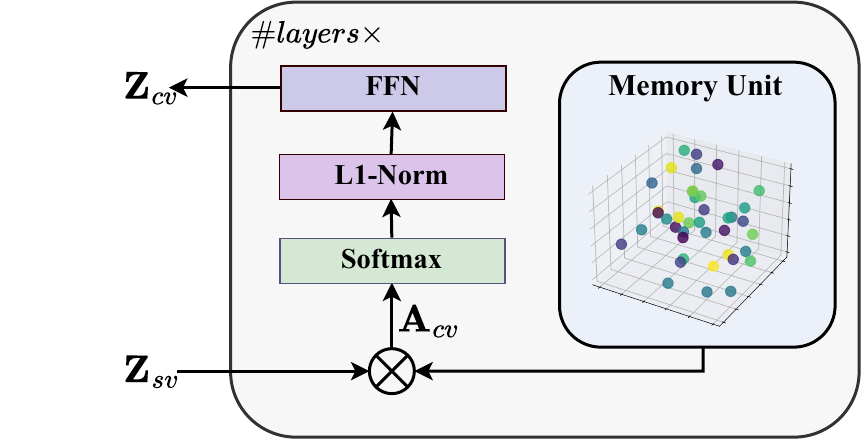}
     \caption{Inter-view attentive feature learning (\moduleB).}
     \label{fig:crossviewafm}
\end{figure}

To avoid these issues, we introduce a learnable \emph{memory unit}~\cite{externalA} to help effectively and efficiently learn the correlations among the regions across views.  
Fig.~\ref{fig:crossviewafm} shows the structure of \moduleB.
The memory unit stores a list of $d_m$ ($d_m = 72$ in the experiments) learned \emph{representative embeddings} in the latent region embedding space (which can be thought of as the cluster centers of the region embeddings). We learn the correlations between each region and the representative embeddings in the memory unit to implicitly learn the correlation between the region and the rest of the regions, since the memory unit serves as a summary of the latent embedding space. As there are limited embeddings in the memory unit, the learning process can be done efficiently. 

\moduleB~reads the embedding matrices $\{\mathbf{Z}_{sv}^j\}_{j=1}^{v}$ from \moduleA\ and concatenates them into one large matrix $\mathbf{Z}_{sv} \in \mathbb{R}^{n \times v \times d}$ for ease of process. 
It computes the correlation coefficients between all regions in $\mathbf{Z}_{sv}$ and the memory unit:
\begin{equation}
\mathbf{A}_{cv} = \mathrm{FFN}(\mathbf{Z}_{sv})
\end{equation}
where $\mathbf{A}_{cv}$ denotes \emph{inter-view region correlation coefficients} between regions. 
$\mathrm{FFN}(\cdot)$ denotes a feedforward neural network, where the weight matrix is in  $\mathbb{R}^{d \times d_m}$. The $d_m$ vectors in this weight matrix are considered as the representative embeddings (of $d$ dimensions). They can be seen as $d_m$ randomly sampled cluster centers in the latent region embedding space. 

Then, the inter-view region correlation coefficients are fed into two normalization layers (i.e., a Softmax layer and an L1-Norm layer) and another FFN to produce embeddings $\mathbf{Z}_{cv}$ embedded with cross-view region correlation coefficients:
\begin{equation}
\mathbf{Z}_{cv} = \mathrm{FFN}(\mathrm{L1{\text -}{Norm}}(\mathrm{Softmax(\mathbf{A}_{cv})}))
\end{equation}
where the weight matrix of the FFN is in $\mathbb{R}^{d_m \times d}$. Here, the Softmax layer is applied on the second dimension (i.e., the view dimension), and the L1-Norm layer is applied on the third dimension (i.e., the embedding dimension). We repeat the computations above for multiple layers to enhance the model learning capacity. The final $\mathbf{Z}_{cv}$ output of \moduleB\ can be decomposed into a list of sub-matrices by the second dimension (the views), i.e., $\mathbf{Z}_{cv}^1, \mathbf{Z}_{cv}^2, ..., \mathbf{Z}_{cv}^v$, where each sub-matrix is in $\mathbb{R}^{n \times d}$. We call these matrices the \emph{inter-view attentive region embedding matrices}.

Finally, we adaptively combine the intra-view attentive region embedding matrices $\{\mathbf{Z}_{sv}^j\}$ and the inter-view attentive region embedding matrices $\{\mathbf{Z}_{cv}^j\}$ with a learnable weight $\beta \in [0,1]$, to form the view-based region embeddings $\{\mathbf{Z}^j\}$:
\begin{equation}
{\mathbf{Z}^j} = \beta{\mathbf{Z}_{sv}^j} + (1-\beta){\mathbf{Z}_{cv}^j}
\end{equation}
\section{EXPERIMENTS}\label{sec:exp}
We perform an empirical study to verify:
 (Q1)~the accuracy of our proposed model~\textbf{\model} on three downstream tasks as compared with the state-of-the-art (SOTA) methods, (Q2)~the general applicability of our proposed dual-feature attentive fusion when incorporated into existing multi-view region representation learning models, (Q3)~the effectiveness of our model components and different input views, (Q4)~the impact of number of regions and population density to our model, and 
(Q5)~the impact of model hyper-parameters. 

\begin{table}[ht]
\centering
\caption{Dataset Statistics}
\label{tab:datasets}
\renewcommand{\arraystretch}{1.2}
\setlength{\tabcolsep}{2pt}
\resizebox{0.95\columnwidth}{!}{
\begin{tabular}{l|r|r|r}
\hlineB{3}
&\textbf{NYC~\cite{nycOpendata}} & \textbf{CHI~\cite{chiOpendata}} & \textbf{SF~\cite{sfOpendata}} \\ \hline \hline
\#regions & 180 & 77 & 175 \\

\#POIs & 24,496 & 57,891 & 28,578 \\

\#POI categories & 26 & 26 & 26 \\

\#land use categories  & 11 & 12 & 23 \\

{\#taxi trips} & 10,953,879 & 3,381,807 & 357,749 \\
(data collection time) &  06/2015 - 07/2015 & 01/2021 - 01/2022 & 05/2008 - 06/2008 \\

{\#crime records} & 35,335 & 18,200  & 48,489 \\
(data collection time) & unknown & 12/2022 - 12/2022 & 01/2022 - 12/2022 \\

{\#check-ins} & 106,902 & 167,232 & 87,750 \\
(data collection time) & 04/2012 - 09/2013 & 04/2012 - 09/2013 & 04/2012 - 09/2013 \\

{\#service calls} & 516,187 & 24,350 & 34,385 \\ 
(data collection time) & 01/2023 - 03/2023 & 12/2022 - 12/2022 & 01/2022 - 12/2022 \\ \hlineB{3}
\end{tabular}
}
\vspace{-1mm}
\end{table}

\subsection{Experimental Settings}\label{sec:exp_settings}
\textbf{Dataset.} The experiments are conducted using real-world data from cities: New York City (\textbf{NYC})~\cite{nycOpendata}, Chicago (\textbf{CHI})~\cite{chiOpendata}, and San Francisco (\textbf{SF})~\cite{sfOpendata}.
Previous works~\cite{MVURE, MGFN, HREP} only used NYC, while CHI and SF are used for the first time in our study, to show the general applicability of our model across different datasets.

Table~\ref{tab:datasets} summarizes the datasets. For each city, we have:

(1) region boundary of each region;

(2) taxi trip records including the pickup and drop-off points of each trip -- we use the numbers of trips from a region to the others as the mobility feature vector of the region;

(3) POIs in each region (from OpenStreetMap~\cite{osm}), along with their  category labels as described in Section~\ref{sec:preliminaries};
 
(4)) zones in each region, along with their land use category labels  as described in Section~\ref{sec:preliminaries} -- there are 11, 12, and 13 categories for NYC, CHI, and SF, respectively;

(5) crime, check-in (from a Foursquare dataset~\cite{checkinData}), and service call records, each with a location and a time -- we count the number of these records in each region, respectively, and predict the counts using the region embeddings learned from the features above, following previous works~\cite{MVURE, MGFN, HREP}. 

\textbf{Prediction model.}
We use three downstream tasks for model evaluation, i.e., 
crime, check-in, and service call count prediction, which are regression tasks. We implement a Lasso regression model (model parameter $\alpha=1$) for each prediction task, following previous works~\cite{MGFN, HREP, MVURE}.

\textbf{Evaluation metrics.} We use three classic evaluation metrics for regression tasks: Mean Absolute Error (\textbf{MAE} = $\frac{\sum_{i=1}^{n} |y_i - \hat{y_i}|}{n}$), Root Mean Square (\textbf{RMSE} = $\sqrt{\frac{\sum_{i=1}^{n} (y_i - \hat{y_i})^{2}}{n}}$), and Coefficient of Determination ($\boldsymbol{R^2} = 1 - \frac{\sum_{i=1}^{n}(y_i - \hat{y_i})^2}{\sum_{i=1}^{n}(y_i - \bar{y})^2}$).

\textbf{Competitors.} We compare with the following models, including the SOTA models RegionDCL \cite{RegionDCL} and HREP \cite{HREP}:

(1)~\textbf{MVURE~\cite{MVURE}} leverages both the intra-region (POI and check-in features) and inter-region features (human mobility features) to construct multi-view graphs, which is followed by multi-view fusion to learn region embeddings. 

(2)~\textbf{MGFN~\cite{MGFN}} leverages human mobility features to construct mobility pattern graphs by clustering mobility graphs based on their spatio-temporal distances. Afterwards, it performs inter-view and intra-view messages passing on the mobility pattern graphs to generate region embeddings.

(3)~\textbf{RegionDCL~\cite{RegionDCL}} leverages building footprints from OpenStreetMap and uses contrastive learning at both the building-group and region levels to learn building-group embeddings. Afterwards, it aggregates all building-group embeddings in a region to generate region embeddings.

(4)~\textbf{HREP~\cite{HREP}} leverages human mobility, POI, and geographic neighbor features to generate region embeddings. Subsequently, it integrates task-based learnable prompt embeddings with the pre-trained region embeddings to customize the embeddings for different downstream tasks.

\begin{table*}[t] \scriptsize
\captionsetup{justification=centering}
\caption{Overall Prediction Accuracy Results (`$\downarrow$' indicates that smaller values are preferred, and `$\uparrow$' indicates that large values are preferred. The best results are in boldface, and the second-best results are underlined. Same for the tables below.)}
\label{tab:overall_results}
\setlength{\tabcolsep}{3pt} 
\renewcommand{\arraystretch}{1.1} 
\resizebox{\textwidth }{!}{
\begin{tabular}{l *{9}{c}}
    \toprule [0.4ex] \\[-3.5ex]

    \multirow{2}{*}{\textbf{\makecell[l]{Taks~1: \\  Check-in}}}& \multicolumn{3}{c}{New York City} & \multicolumn{3}{c}{Chicago} & \multicolumn{3}{c}{San Francisco}\\
    
    \cmidrule(lr){2-4} \cmidrule(lr){5-7} \cmidrule(lr){8-10}
    &MAE $\downarrow$&RMSE $\downarrow$&$R^{2} \uparrow$ &MAE $\downarrow$ &RMSE $\downarrow$&$R^{2} \uparrow$ &MAE $\downarrow$&RMSE $\downarrow$&$R^{2} \uparrow$ \\
    \midrule
    MVURE~\cite{MVURE} & 306.7 $\pm$ 8.20 & 499.6 $\pm$ 12.9 & 0.627 $\pm$ 0.019 & 1693 $\pm$ \, 74 & 3171 $\pm$ 128 & 0.656 $\pm$ 0.029  & 346.8 $\pm$ 8.7 & 659.3 $\pm$ 15.7 & 0.562 $\pm$ 0.021  \\
    MGFN \cite{MGFN} & 292.6 $\pm$ 17.1 & 451.8 $\pm$ 28.1 & 0.690 $\pm$ 0.040  & \underline{1281 $\pm$ \, 41} & \underline{2276 $\pm$ \, 86} & \underline{0.817 $\pm$ 0.011}  & \underline{310.8 $\pm$ 9.1} & \underline{542.1 $\pm$ 17.6} & \underline{0.708 $\pm$ 0.010}  \\
    RegionDCL~\cite{RegionDCL} & 371.2 $\pm$ 10.3 & 495.5 $\pm$ 15.9 & 0.471 $\pm$ 0.023  & 2427 $\pm$ 123 & 4184 $\pm$ 136 & 0.402 $\pm$ 0.042  & 398.8 $\pm$ 9.9 & 748.1 $\pm$ 17.8 & 0.437 $\pm$ 0.024  \\
    HREP~\cite{HREP} & \underline{276.3 $\pm$ 11.7} & \underline{448.2 $\pm$ 17.1} & \underline{0.703 $\pm$ 0.021}  & 1679 $\pm$ \, 71 & 3135 $\pm$ \, 79 & 0.664 $\pm$ 0.017  & 330.9 $\pm$ 9.3 & 606.7 $\pm$ 25.8 & 0.629 $\pm$ 0.032 \\
    \midrule 
    \textbf{\model} & \textbf{202.8 $\pm$ 7.2} & \textbf{322.8 $\pm$ 12.6} & \textbf{0.844 $\pm$ 0.012}  & \textbf{\,  929 $\pm$ \, 62} & \textbf{1947 $\pm$ \, 75} & \textbf{0.870 $\pm$ 0.010}  & \textbf{233.1 $\pm$ 9.5} & \textbf{429.6 $\pm$ 28.1} & \textbf{0.813 $\pm$ 0.024} \\
    \midrule 
    \textbf{Improvement} &\textbf{26.6\%} &\textbf{28.0\%} &\textbf{20.6\%} &\textbf{27.4\%} &\textbf{14.5\%} &\textbf{6.5\%} &\textbf{25\%} &\textbf{20.7\%} &\textbf{14.8\%} \\

    \bottomrule \toprule [0.4ex] \\[-3.5ex]

    \multirow{2}{*}{\textbf{\makecell[l]{Task~2: \\ Crime}}}& \multicolumn{3}{c}{New York City} & \multicolumn{3}{c}{Chicago} & \multicolumn{3}{c}{San Francisco}\\
    \cmidrule(lr){2-4} \cmidrule(lr){5-7} \cmidrule(lr){8-10}
    &MAE $\downarrow$&RMSE $\downarrow$&$R^{2} \uparrow$ &MAE $\downarrow$ &RMSE $\downarrow$&$R^{2} \uparrow$ &MAE $\downarrow$&RMSE $\downarrow$&$R^{2} \uparrow$ \\
    \midrule
    MVURE~\cite{MVURE} & 67.9 $\pm$ 1.1 & \, 93.8 $\pm$ 1.9 & 0.591 $\pm$ 0.016 & 100.4 $\pm$ 6.6 & 129.2 $\pm$ 7.3 & 0.461 $\pm$ 0.062  & 130.3 $\pm$ 1.7 & 201.7 $\pm$ 3.2 & 0.594 $\pm$ 0.013  \\
    MGFN~\cite{MGFN} & 70.2 $\pm$ 2.3 & \, 89.6 $\pm$ 2.5 & 0.630 $\pm$ 0.020  & 107.4 $\pm$ 5.4 & 137.9 $\pm$ 5.2 & 0.386 $\pm$ 0.047  & 128.4 $\pm$ 3.3 & 199.9 $\pm$ 4.3 & 0.601 $\pm$ 0.017  \\
    RegionDCL~\cite{RegionDCL} & 98.7 $\pm$ 3.1 & 127.9 $\pm$ 5.2 & 0.251 $\pm$ 0.026  & 121.7 $\pm$ 4.8 & 159.6 $\pm$ 6.3 & 0.179 $\pm$ 0.053  & 156.3 $\pm$ 2.1 & 242.3 $\pm$ 4.6 & 0.413 $\pm$ 0.021  \\
    HREP~\cite{HREP} & \underline{62.8 $\pm$ 2.1} & \underline{\, 83.1 $\pm$ 2.3} & \underline{0.680 $\pm$ 0.014}  & \underline{\, 88.3 $\pm$ 6.4} & \underline{114.4 $\pm$ 5.5} & \underline{0.578 $\pm$ 0.041}  & \underline{124.4 $\pm$ 2.3} & \underline{196.9 $\pm$ 3.9} & \underline{0.612 $\pm$ 0.014} \\
    \midrule 
    \textbf{\model} & \textbf{56.1 $\pm$ 1.3} & \textbf{\, 76.1 $\pm$ 2.2} & \textbf{0.734 $\pm$ 0.015}  & \textbf{\, 77.8 $\pm$ 3.6} & \textbf{107.1 $\pm$ 5.4} & \textbf{0.631 $\pm$ 0.036}  & \textbf{101.5 $\pm$ 3.3} & \textbf{178.4 $\pm$ 3.6} & \textbf{0.682 $\pm$ 0.013} \\
    \midrule 
    \textbf{Improvement} & \textbf{10.8\%} & \textbf{8.4\%} &\textbf{7.8\%} &\textbf{11.9\%} &\textbf{6.4\%} & \textbf{9.2\%}  &\textbf{18.4\%} &\textbf{9.4\%} &\textbf{11.4\%} \\
     \bottomrule \toprule [0.4ex] \\[-3.5ex]

    \multirow{2}{*}{\textbf{\makecell[l]{Task~3: \\ Service call}}}& \multicolumn{3}{c}{New York City} & \multicolumn{3}{c}{Chicago} & \multicolumn{3}{c}{San Francisco}\\
    \cmidrule(lr){2-4} \cmidrule(lr){5-7} \cmidrule(lr){8-10}
    &MAE $\downarrow$&RMSE $\downarrow$&$R^{2} \uparrow$ &MAE $\downarrow$ &RMSE $\downarrow$&$R^{2} \uparrow$ &MAE $\downarrow$&RMSE $\downarrow$&$R^{2} \uparrow$ \\
    \midrule
    MVURE~\cite{MVURE} & \underline{1428 $\pm$ 33} & 2180 $\pm$ \, 46 & 0.367 $\pm$ 0.027 & 190.3 $\pm$ \, 9.8 & 266.9 $\pm$ 12.1 & 0.441 $\pm$ 0.050  & \underline{102.1 $\pm$ 4.8} & \underline{164.7 $\pm$ 2.7} & \underline{0.479 $\pm$ 0.017}  \\
    MGFN~\cite{MGFN} & 1554 $\pm$ 81 & 2286 $\pm$ 115 & 0.303 $\pm$ 0.069  & 208.2 $\pm$ 11.3 & 293.4 $\pm$ 16.6 & 0.329 $\pm$ 0.077  & 102.8 $\pm$ 2.2 & 166.3 $\pm$ 2.5 & 0.468 $\pm$ 0.021  \\
     RegionDCL~\cite{RegionDCL} & 1783 $\pm$ 21 & 2597 $\pm$ \, 38 & 0.103 $\pm$ 0.026  & 195.7 $\pm$ \, 7.6 & 272.1 $\pm$ 10.1 & 0.445 $\pm$ 0.041  & 116.6 $\pm$ 2.3 & 196.7 $\pm$ 3.2 & 0.256 $\pm$ 0.024  \\
    HREP~\cite{HREP} & 1430 $\pm$ 29 & \underline{2286 $\pm$ \, 34} & \underline{0.398 $\pm$ 0.021}  & \underline{185.7 $\pm$ \, 6.1} & \underline{262.2 $\pm$ 10.8} & \underline{0.468 $\pm$ 0.022}  & 103.4 $\pm$ 3.2 & 167.4 $\pm$ 4.6 & 0.461 $\pm$ 0.029 \\
    \midrule 
    \textbf{\model} & \textbf{1273 $\pm$ 20} & \textbf{1951 $\pm$ \, 27} & \textbf{0.493 $\pm$ 0.014}  & \textbf{159.3 $\pm$ 13.9} & \textbf{222.0 $\pm$ 18.9} & \textbf{0.613 $\pm$ 0.067}  & \textbf{\, 81.5 $\pm$ 2.5} & \textbf{142.1 $\pm$ 3.2} & \textbf{0.612 $\pm$ 0.018} \\
    \midrule 
    \textbf{Improvement} &\textbf{10.9\%} &\textbf{8.3\%} &\textbf{23.9\%} & \textbf{14.2\%} &\textbf{15.3\%} &\textbf{31.0\%} &\textbf{20.2\%} &\textbf{13.7\%} &\textbf{27.8\%} \\

    \bottomrule
\end{tabular}
}
\vspace{-4mm}
\end{table*}

\textbf{Hyperparameter settings.}
For the competitor models, we use parameter settings recommended in their papers as much as possible. Special settings have been made on CHI, to prevent model overfitting on this dataset which has fewer regions. For MGFN, we employ a 1-layer (instead of the default 3-layer) mobility pattern joint learning module on CHI. For MVURE and HREP, we reduce the number of layers in their GNN modules from 3 to 2 on CHI. RegionDCL does not need a special setting on CHI because the training process is determined by the number of buildings and building groups, rather than the number of regions.

For our model \model, the number of layers in the \moduleA\ module is 3 for NYC and SF and 1 for CHI, respectively. In the \moduleB\ module, the number of layers is 3 for NYC and 2 for CHI and SF, respectively. The number of layers in the \moduleD\ module is 3 for all three datasets. These parameter values are set by a grid search. 
We train our model for 2,500 epochs in full batches, using Adam optimization with a learning rate of 0.0005. 

The region embedding dimensionality $d$ is set as 144 for~\model\ following an SOTA model HREP~\cite{HREP}, and 96, 96, and 64 for MVURE, MGFN, and RegionDCL, respectively, as suggested by their original papers. Our experimental results in Section~\ref{sec:parameter} also show that these dimensionality values are optimal for the respective models.

We note that MVURE is designed to take check-in records as part of its input. We follow this setting and use check-in records to train MVURE even for the check-in prediction task. The training and testing data come from non-overlapping time periods, i.e.,  MVURE has not seen the testing data at training.

\subsection{Overall Results}
We evaluate the quality of the learned region embeddings through using them in three downstream prediction tasks.
We first learn region embeddings using each model on data from each city, respectively. We subsequently feed the learned region embeddings into a Lasso regression model for each prediction task, employing ten-fold cross-validation (because the number of regions in each dataset is relatively small), and report in Table~\ref{tab:overall_results} the average prediction accuracy results. 

\subsubsection{Task 1: Check-in Count Prediction}
Predicting check-in counts for regions can provide guidance for urban planning and business decision-making (e.g., movement tracking~\cite{10.1007/s00778-014-0358-x,WANG20141}), based on people's locations and interests.

We make the following observations from Table~\ref{tab:overall_results} (Task~1):

(1)~Our model \model\ outperforms all competitors across all datasets with up to 28\% improvement in RMSE. Our model not only considers various types of features of each region but also learns the correlation among the features of different regions both within a single view and across multiple views, which captures the feature patterns to a full extent. Simultaneously, we effectively fuse the patterns to extract their joint impact, leading to highly effective embeddings for region-based prediction tasks.

(2)~MGFN, which only considers one input feature (i.e., human mobility), outperforms (up to 20\% in terms of $R^2$) the baseline methods MVURE and HREP that use human mobility, POI, geographic neighbor, and check-in features, on CHI and SF datasets, while it performs worse on NYC. There are three main reasons: (i)~human mobility is intrinsically linked to the check-in count of each region. (ii)~MGFN explores the complex mobility patterns from the human mobility data, allowing it to learn the relationships among regions from such patterns. (iii)~the human mobility data of NYC is noisy, which negatively impacts the learning effectiveness of MGFN. In this case, the limitations of relying on a single data feature become evident, emphasizing the necessity of a multi-view approach.

(3)~HREP outperforms MVURE across all three datasets, as it introduces a relational embedding to capture different correlations among regions and the importance of the correlations.   
It also incorporates prompt learning to add task-specific
prompt embeddings for different downstream tasks.

(4)~RegionDCL performs the worst, even though it is one of the SOTA models. It only utilizes building footprints to learn region embeddings (its original targeted downstream tasks are population and land use predictions~\cite{RegionDCL}), which do not exhibit a strong correlation with the check-in counts. Furthermore, distinguishing the functionality of regions by buildings can be challenging in cities such as NYC, where buildings predominantly take on a rectangular shape, irrespective of whether they are situated in industrial or residential areas. In addition, RegionDCL is based on contrastive learning, which is biased to generate similar embeddings for regions nearby.

\subsubsection{Task 2: Crime Prediction}
Predicting crime counts in different regions is a valuable tool for law enforcement and community safety, as it allows for a more proactive and strategic approach to crime prevention and reduction.

Table~\ref{tab:overall_results} (Task 2) presents model performance for such a task. 
Our model~\model\ again outperforms the best competitor model HREP across all datasets (7.8\% on NYC, 9.2\% on CHI, and 11.4\% on SF in $R^2$), for our more effective multi-view fusion techniques to capture inter-region relationships.

The baseline models MVURE and HREP that take multiple types of input features are now also better than those taking only a single type of features, i.e., MGFN and RegionDCL. Crime counts are impacted by multiple factors. Regions with high crime counts typically have a high level of human mobility, many entertainment venues such as bars and clubs, and are proximity to public transportation. These factors cannot be reflected by only a single type of region features.

\subsubsection{Task 3: Service Call Prediction}
Predicting service calls in each region is a valuable tool for service providers, allowing proactive optimization of facility and service deployments.
As Table~\ref{tab:overall_results} shows, our model~\model\ achieves consistent and substantial improvements over all competitors, with an advantage over the best baseline by 23.9\% on NYC, 31\% on CHI, and 27.8\% on SF in $R^2$, respectively, further confirming the robustness of the embeddings learned by~\model.

Like in the crime count prediction task, the 
baseline models MVURE and HREP which consider multiple types of input features, once again outperform MGFN and RegionDCL.

Model performance on NYC is, in general, lower than that on CHI and SF. This is because NYC contains about 400 categories of service calls, including noise complaints, graffiti cleanup, etc., while SF and CHI only have 67 and 104, respectively. It is more challenging to generate region embeddings that correlate to all these different types of calls.

\begin{table*}[htbp] \scriptsize
\captionsetup{justification=centering}
\caption{Prediction Accuracy Results When Powering Existing Models with Our DAFusion  Module (NYC)}
\label{tab:DualAFM_exp}
\centering
\setlength{\tabcolsep}{3pt} 
\renewcommand{\arraystretch}{1.1} 
\resizebox{\textwidth }{!}{
\begin{tabular}{l *{9}{c}}
    \hlineB{3}
    \multirow{2}{*}{\textbf{New York City}} & \multicolumn{3}{c}{Check-in Prediction} & \multicolumn{3}{c}{Crime Prediction} & \multicolumn{3}{c}{Service Call Prediction}\\
    \cmidrule(lr){2-4} \cmidrule(lr){5-7} \cmidrule(lr){8-10}
     \\[-3ex]
    &MAE $\downarrow$&RMSE $\downarrow$&$R^{2} \uparrow$ &MAE $\downarrow$ &RMSE $\downarrow$&$R^{2} \uparrow$ &MAE $\downarrow$&RMSE $\downarrow$&$R^{2} \uparrow$ \\ 
    \toprule
    MGFN & 292.6 $\pm$ 17.1 & 451.8 $\pm$ 28.1 & 0.690 $\pm$ 0.040 & 70.2 $\pm$ 2.3 &89.6 $\pm$ 2.5 & 0.630 $\pm$ 0.020 & 1553.8 $\pm$ 80 &2286 $\pm$ 115 & 0.303 $\pm$ 0.069 \\

    \textbf{MGFN-DAFusion} &\textbf{222.0 $\pm$ \, 5.7} &\textbf{363.6 $\pm$ \, 8.5} &\textbf{0.802 $\pm$ 0.009} &\textbf{59.8 $\pm$ 1.2} &\textbf{81.1 $\pm$ 2.4} &\textbf{0.699 $\pm$ 0.018}  &\textbf{1375.7 $\pm$ 16} &\textbf{2101 $\pm$ \, 29} &\textbf{0.412 $\pm$ 0.016} \\
    \midrule 
    \textbf{Improvement} &\textbf{24.2\%} &\textbf{19.5\%} &\textbf{16.2\%} &\textbf{14.8\%} &\textbf{9.5\%} &\textbf{11.0\%} &\textbf{11.5\%} &\textbf{8.1\%} &\textbf{36.0\%} \\
     \midrule \midrule
    MVURE & 306.7 $\pm$ 8.2 & 499.6 $\pm$ 12.9 & 0.627 $\pm$ 0.019 & 67.9 $\pm$ 1.1 &93.8 $\pm$ 1.9 & 0.591 $\pm$ 0.016  & 1428 $\pm$ 33 &2180 $\pm$ 46 & 0.367 $\pm$ 0.027 \\

    \textbf{MVURE-DAFusion} &\textbf{275.8 $\pm$ 5.9} &\textbf{438.5 $\pm$ 14.4} &\textbf{0.712 $\pm$ 0.018} &\textbf{62.6 $\pm$ 1.5} &\textbf{85.9 $\pm$ 1.9} &\textbf{0.663 $\pm$ 0.015} &\textbf{1317 $\pm$ 16} &\textbf{2049 $\pm$ 33} &\textbf{0.441 $\pm$ 0.018} \\
    \midrule 
    \textbf{Improvement} &\textbf{10.1\%} &\textbf{12.2\%} &\textbf{13.6\%} &\textbf{7.8\%} &\textbf{8.4\%} &\textbf{12.4\%} &\textbf{7.8\%} &\textbf{6.0\%} &\textbf{20.2\%} \\
     \midrule \midrule
    HREF & 276.3 $\pm$ 11.7 & 448.2 $\pm$ 16.9 & 0.703 $\pm$ 0.022 & 62.8 $\pm$ 2.1 &83.1 $\pm$ 2.3 & 0.681 $\pm$ 0.014  & 1430 $\pm$ 29 & 2127 $\pm$ 34 & 0.398 $\pm$ 0.019 \\

    \textbf{HREF-DAFusion} &\textbf{224.6 $\pm$ \, 8.5} &\textbf{360.0 $\pm$ 10.3} &\textbf{0.806 $\pm$ 0.011} &\textbf{59.2 $\pm$ 1.6} &\textbf{78.6 $\pm$ 1.3} &\textbf{0.717 $\pm$ 0.011} &\textbf{1325 $\pm$ 19} &\textbf{2031 $\pm$ 29} &\textbf{0.451 $\pm$ 0.015} \\
    \midrule 
    \textbf{Improvement} &\textbf{18.7\%} &\textbf{19.7\%} &\textbf{15.1\%} &\textbf{5.7\%} &\textbf{5.4\%} &\textbf{5.3\%} &\textbf{7.3\%} &\textbf{4.5\%} &\textbf{13.3\%} \\
    \bottomrule
\end{tabular}
}
\vspace{-3mm}
\end{table*}

\begin{table}[ht]
\centering
\caption{Embedding Learning and Testing Time (seconds)}
\label{tab:computation_time}
\setlength{\tabcolsep}{4pt}
\renewcommand{\arraystretch}{1.1}
\resizebox{1\columnwidth}{!}{
\begin{tabular}{l|c|ccc|ccc}
\hlineB{3}

&& \multicolumn{3}{c|}{\textbf{Embedding Learning}} & \multicolumn{3}{c}{\textbf{Downstream Task}} \\ \hline 
& & \textbf{NYC} & \textbf{CHI} &\textbf{SF} & \textbf{NYC} & \textbf{CHI} &\textbf{SF} \\ \hline \hline
\multirow{2}{*}{MVURE}& CPU &557 &153 &518  &0.016 &\textbf{0.049 }&0.028 \\ 
  & GPU  & \textbf{35} & \textbf{15}& \textbf{34} & 0.023& 0.053&0.026\\ 
\hline

\multirow{2}{*}{MGFN}&CPU &1,452&460&1,388 &0.017&0.054&0.032\\ 
&GPU &92 &123 &47 &0.019 &0.061 &0.029\\ 
\hline

\multirow{2}{*}{RegionDCL}& CPU &2,004&17,435&5,748 & \textbf{0.014} &0.051 &0.026\\ 
& GPU &149 &1,779 &324 & 0.017 &0.054 &\textbf{0.023}\\ 
\hline

\multirow{2}{*}{HREP}&CPU &302 &136 &267 &171 &325 &152\\ 
&GPU &51 &45 &51 &92 &146 &91\\ 
\hline

\multirow{2}{*}{\model}& CPU &934&346&856 &0.019&0.062&0.037\\ 
& GPU &79 &51 &78 &0.022&0.061&0.028\\ 
\hlineB{3}
\end{tabular}
}
\end{table}
\vspace{1mm}

\textbf{Model running time.}  Table~\ref{tab:computation_time} reports the embedding learning and downstream task running times (including regression model learning and inference, and they have the same input and output size irrespective to the downstream tasks). All models are trained and tested on a machine equipped with a 16-core Intel i7 2.6 GHz CPU and 16 GB memory, and a machine equipped with an NVIDIA Tesla V100 GPU and 64 GB of memory, to test the efficiency of the models on CPU and on GPU, respectively (same for the other experiments).

\model\ takes extra time to learn the embeddings because of its extra attention computation and fusion steps. However, we see that its embedding learning time is still on the same order of magnitude as that of the fastest model MVURE, on both CPU and GPU. It can be trained on GPU in under 80 seconds, making it highly practical in terms of training costs.

MVURE and HREP simply aggregate region embeddings from all views, which are faster to train, but this efficiency comes at the cost of significantly higher prediction errors with their learned embeddings, as shown above.

All models share the same prediction model for the downstream tasks, and hence their downstream task running times are very similar (except for HREP). The slight running time differences are due to the difference in the embedding dimensionality. The running time difference between GPU and CPU is insignificant, as the prediction model has a simple structure  that limits the potential benefits from GPU parallelization.  
HREP is much slower because it requires a prompt embedding learning step for each downstream task. 

\subsection{Applicability of Dual-feature Attentive Fusion}\label{sec:exp_dualfusion}

To show the general applicability of our dual-feature attentive fusion module (\fusion), we integrate it with the three baseline models that compute multiple views and require an embedding fusion, i.e., MVURE, MGFN, and HREP. We denote the resulting models as \textbf{MVURE-\fusion}, \textbf{MGFN-\fusion}, and \textbf{HREP-\fusion}, respectively. 

We run experiments with the three datasets and three downstream tasks on these new model variants, with 10-fold cross-validation as done above. We show the results on NYC in Table~\ref{tab:DualAFM_exp} for conciseness, as the results on the other two datasets have a similar comparative pattern. We see that \fusion\ also enhances the effectiveness of the existing multi-view based models. Compared with the vanilla models, the variants powered by \fusion\ achieve significant improvements, i.e., up to 16.2\%, 12.4\%, and 36\% in ${R^{2}}$, across the three downstream tasks. These results confirm the effectiveness and applicability of our \fusion\ module. The improvement on MGFN, in particular, indicates that \fusion\ can also better integrate information from multiple graphs even when the graphs are constructed from a single type of features.

By further comparing the results in Tables~\ref{tab:overall_results} and~\ref{tab:DualAFM_exp}, we note that \model\ still outperforms the baseline models even when they are powered by \fusion, which confirms the contribution of our hybrid attentive feature learning module to the overall model accuracy. 

\vspace{-1mm}
\subsection{Ablation Study}\label{sec:exp_ablation}
We use the following model variants to study the effectiveness of our model components:

(1)~\textbf{\model-w/o-D$+$:} We replace \fusion\ with a simple element-wise sum of embeddings from different views.

(2)~\textbf{\model-w/o-D$\|$:} We replace \fusion\ with a simple concatenation of the embeddings from different views, followed by an MLP layer to reduce the dimensionality.

(3)~\textbf{\model-w/o-C:} We replace \moduleB~with the vanilla self-attention.

(4)~\textbf{\model-w/o-S:} We replace \moduleA~with the vanilla self-attention.

We again repeat the above experiments with these model variants and report the results in Table~\ref{tab:ablation_study}. We only show the $R^2$ results, as the comparative performance in MAE and RMSE is similar (same below). Our full model \model\ significantly outperforms all variants. This highlights the contribution of all model components to the overall effectiveness of \model. We make the following further observations:

\vspace{1mm}
\begin{table}[htbp] \scriptsize
\caption{Ablation Study Results (NYC)}
\vspace{-2mm}
\label{tab:ablation_study}
\begin{center}
\setlength{\tabcolsep}{3pt}
\renewcommand{\arraystretch}{1.1}
\resizebox{\columnwidth }{!}{
\begin{tabular}{l *{3}{c}}
    \hlineB{3}
    Prediction Task& Check-in & Crime & Service Call\\

    \hline & \\[-2ex]
    &$R^{2} \uparrow$ &$R^{2} \uparrow$ &$R^{2} \uparrow$ \\ 
    \hline & \\[-2ex]
    
    \model-w/o-D$+$ & 0.803 $\pm$ 0.008  & 0.686 $\pm$ 0.015 & 0.459 $\pm$ 0.005 \\
    
    \model-w/o-D$\|$ & 0.816 $\pm$ 0.007 & 0.699 $\pm$ 0.009  &0.468 $\pm$ 0.006 \\

    \model-w/o-C & 0.832 $\pm$ 0.004 & 0.696 $\pm$ 0.016 & 0.462 $\pm$ 0.011 \\

    \model-w/o-S & 0.838 $\pm$ 0.005 & 0.725 $\pm$ 0.017  & 0.482 $\pm$ 0.008 \\

    \hline & \\[-2ex]
    \textbf{\model} &\textbf{0.844 $\pm$ 0.012} &\textbf{0.734 $\pm$ 0.015} &\textbf{0.493 $\pm$ 0.014} \\
    \hlineB{2} & \\[-2ex]
\end{tabular}
}
\end{center}
\end{table}

(1) \fusion\ contributes the most model performance gains. The variants with \fusion, i.e., \model-w/o-C and \model-w/o-S, outperform those without \fusion, i.e.,  \model-w/o-D$+$ and \model-w/o-D$\|$.
This is because the region-aware fusion enables the learning of correlations between regions in the fused representation, leading to more accurate embedding-based prediction results. 

(2) \model-w/o-S performs better than  \model-w/o-C, which indicates that it is important to capture the correlation information between all regions across all views.

\vspace{-1mm}
\subsection{Impact of Different Input Views}\label{sec:view}

\begin{figure}[h]
\vspace{-2mm}
     \centering
     \includegraphics[width = 0.45 \textwidth]{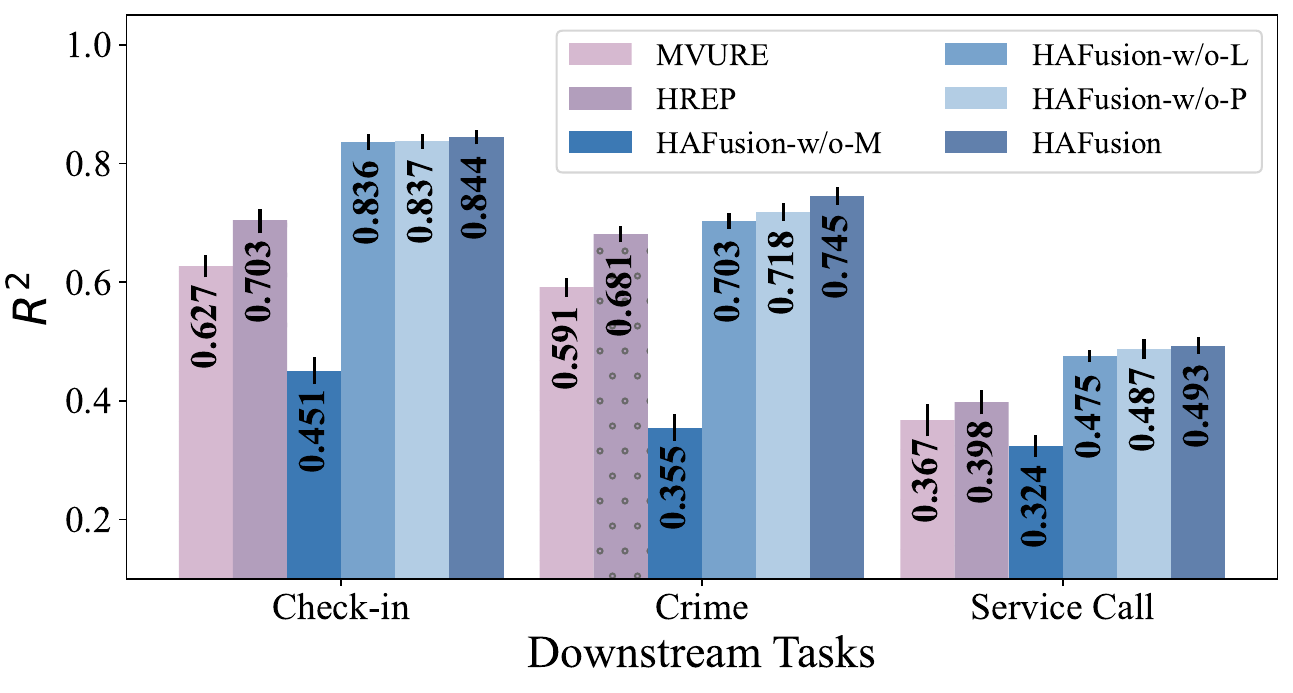}
     \caption{Impact of different input views (NYC).}
     \label{fig:input_features}
\end{figure}

Our model \model~utilizes mobility, POI, and land use views as input in region representation learning. Next, we excluded each of the mobility, POI, and land use views, forming three variants of our model: \textbf{\model-w/o-M}, \textbf{\model-w/o-P}, and \textbf{\model-w/o-L}, respectively. We compare them with the full model \model~to assess the importance of each view. 
We also include MVURE and HREP in this set of experiments, which use human mobility and POI features like \model~does but not the land use view. By comparing with them, we show that \model\ can learn more informative embeddings even without the extra land use view.

The results, as plotted in Fig.~\ref{fig:input_features}, show that the mobility view contributes the most to the overall model performance (i.e., \model-w/o-M performs the worst), as human movement
reflects the correlation between regions, which is particularly relevant to the check-in and crime prediction tasks. 
The land use view, which is used for the first time for urban region representation learning in our work, contributes the second most (i.e., \model-w/o-L is worse than \model-w/o-P), emphasizing the importance of the feature. \model-w/o-L also outperforms MVURE and HREP across all downstream tasks, e.g., by 20\% and 10\% in $R^2$ for check-in prediction, respectively, again demonstrating the effectiveness of our model to better capture region features from the input.

% \vspace{-2mm}
\subsection{Impact of Number of Regions} \label{sec:data_scalability}

\begin{figure}[htbp]
\vspace{-2mm}
\begin{minipage}[t]{0.45\columnwidth}
\captionsetup{font=footnotesize}
  \includegraphics[width=\linewidth]{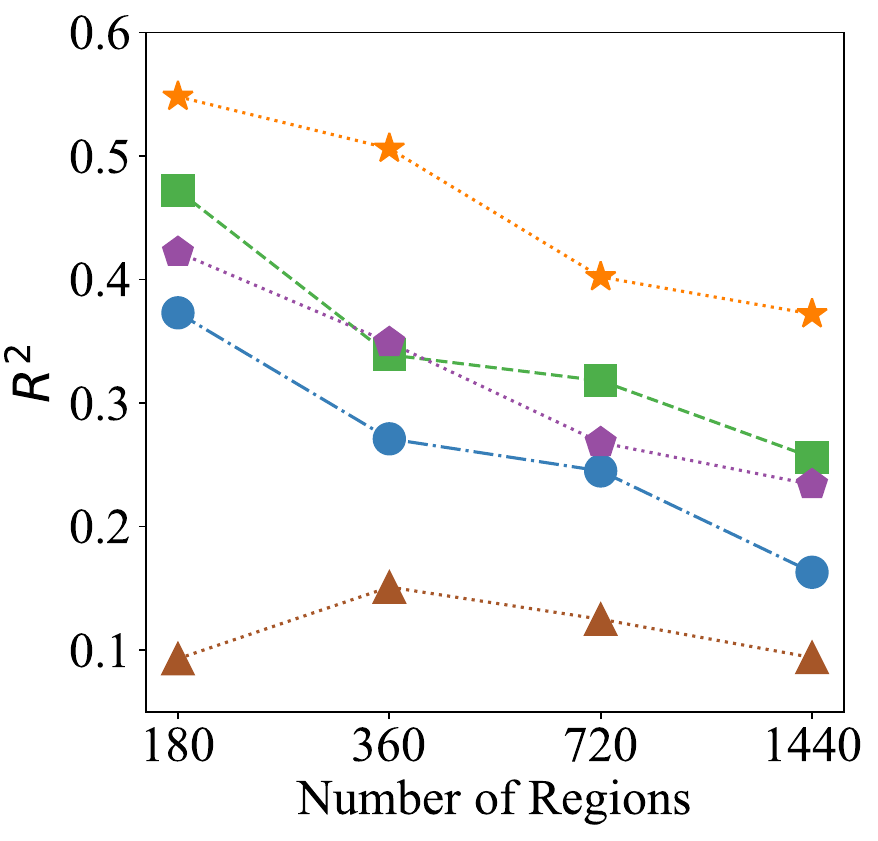}
  \caption*{(a) Checkin prediction accuracy}
\end{minipage}\hfill 
\begin{minipage}[t]{0.45\columnwidth}
\captionsetup{font=footnotesize}
  \includegraphics[width=\linewidth]{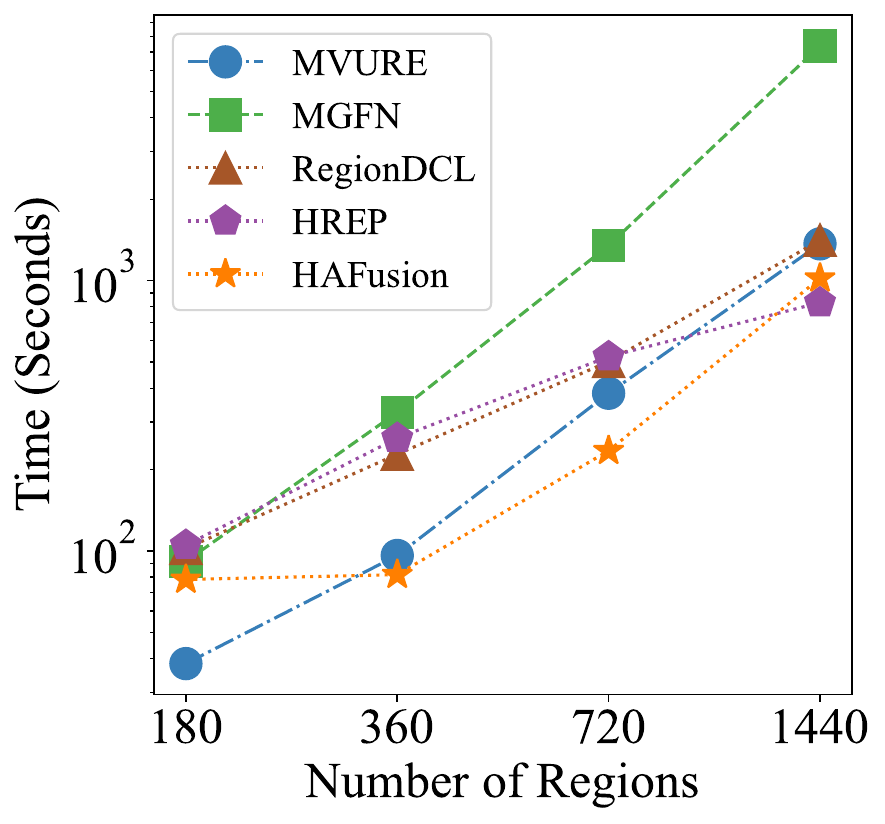}
  \caption*{(b) Overall running time (GPU)}
\end{minipage}
\caption{Impact of number of regions.}
\label{fig:scalarbility}
\end{figure}

We verify the scalability of \model~by varying the number of regions.
We expand the NYC dataset (which originally contains regions from Manhattan) to include regions from Queens and Brooklyn. We start from an urban region in Brooklyn and collect regions in a breath-first manner to construct datasets of 180, 360, 720, and 1,440 regions, respectively, such that the larger datasets are supersets of the smaller ones (the baseline works have used only up to 180 regions).

Fig.~\ref{fig:scalarbility} shows the prediction accuracy using the learned embeddings, as well as the overall running time, which includes the embedding learning
and downstream task running times (regression model learning and inference times). All models suffer in accuracy when the number
of regions grows, because more regions with sparse input features are 
included. Importantly, our model HAFusion consistently outperforms
all competitors in check-in count prediction accuracy, achieving approximately a 10\% improvement over the second-best model across all region settings (similar results are observed on the other prediction tasks).

Meanwhile, the overall running time of \model~is also the lowest, except when there are 180 and 1,440 regions, where \model~is only slower than MVURE and HREP, respectively. The core components of \model~are attention networks, which can be efficiently parallelized on GPUs, and this explains for the low running time.

MVURE utilizes self-attention, which computes in $O(d\cdot N^2)$ time, whereas \model~utilizes external attention, which computes in  $O(d\cdot d_{m} \cdot N)$ time. Recall that $d$ and $d_m$ represent the embedding dimensionality and the number of representative embeddings (cf.~Section~\ref{sec:hybridafm}), respectively, while $N$ denotes the number of regions. Thus, the embedding learning time (and hence the overall running time) of MVURE grows with $N$ faster than that of \model~does.
HREP scales well to the number of regions due to its simple model structure. Its overall running time is dominated by its extra prompt embedding learning time for each downstream task. This extra time is amortized as $N$ grows, and hence the overall running time of HREP grows the slowest, at the cost of producing region embeddings with a lower accuracy.

\subsection{Impact of Population Density}
\label{sec:data_density}

\begin{figure}[ht]
    \vspace{-1mm}
     \centering
     \includegraphics[width = 0.45 \textwidth]{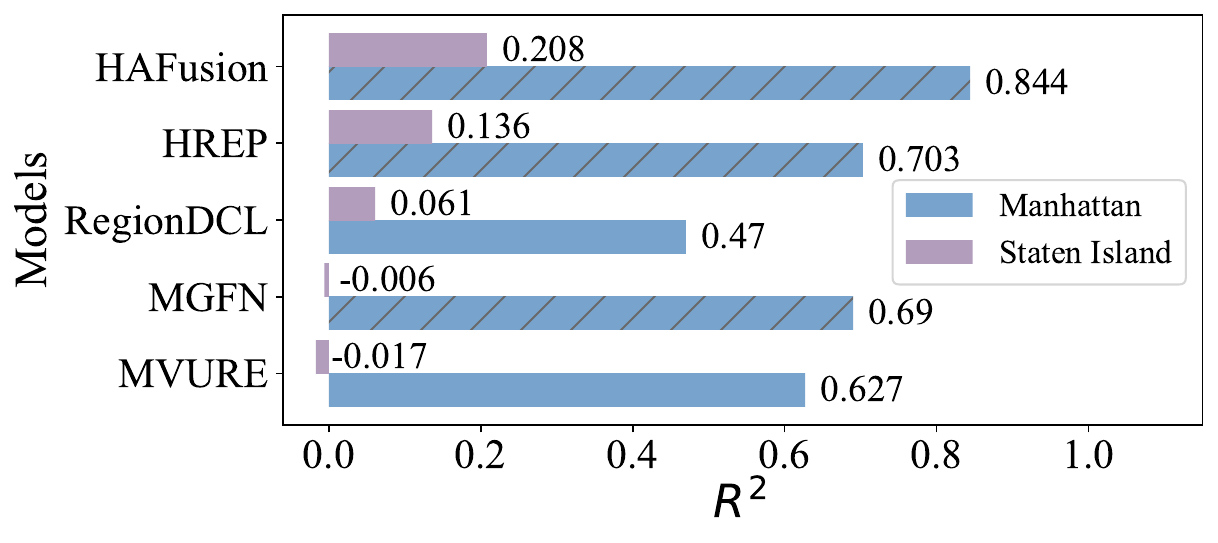}
     \vspace{-1mm}
     \caption{Impact of population density.}
     \label{fig:data_density}
\end{figure}

We further test the models with regions in Staten Island, which
have distinctly different urban characteristics from regions in Manhattan used above. 
Manhattan is the most densely populated and geographically smallest of the five boroughs of NYC. In contrast, Staten Island is the least densely
populated and most suburban borough in NYC. We evaluate the models over regions in these two areas and show the results of the check-in count prediction task (similar results are observed on the other prediction tasks).

As Fig.~\ref{fig:data_density} shows, \model\ outperforms all competitors over both crowded and less crowded regions. 
Existing models suffer heavily over less crowded regions due to limited regional features, particularly mobility features. For example, Manhattan has 10,953,879 records in a single month, whereas Staten Island only has hundreds. MGFN reports the largest drop in accuracy, because it relies solely on mobility features. This again highlights the importance of utilizing multiple types of features, as done by \model.

\subsection{Impact of Parameter Values}\label{sec:parameter}
Next, we study model sensitivity to two key hyper-parameters: the dimensionality of the region embeddings ($d$) and the number of \moduleD~layers in \fusion\ ($\#layers$). 
We note that there are also multiple layers in \moduleA\ and \moduleB. The numbers of those layers are set with grid search as mentioned earlier, and the experimental results are omitted for conciseness.

 \begin{figure}[ht]
\begin{minipage}[t]{0.32\columnwidth}
\captionsetup{font=footnotesize,labelfont=footnotesize}
  \includegraphics[width=\linewidth]{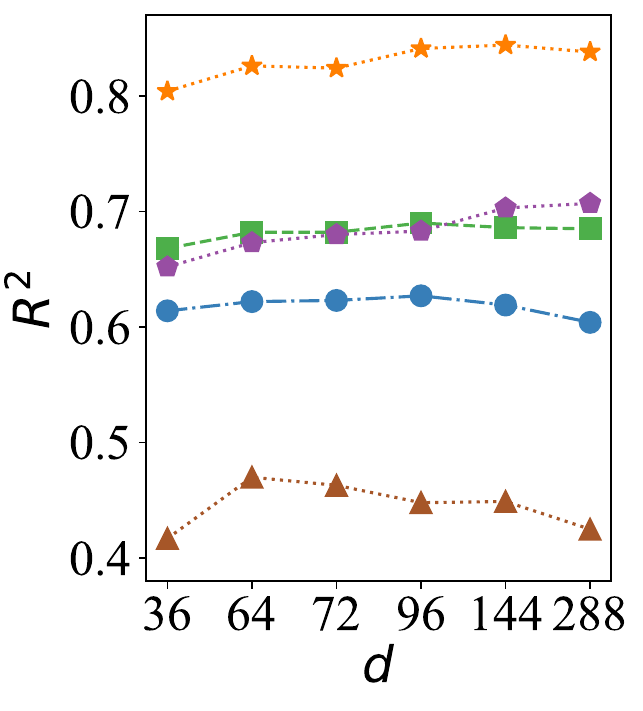}
  \vspace{-5mm}
  \caption*{(a) Check-in}
\end{minipage} 
\begin{minipage}[t]{0.32\columnwidth}
\captionsetup{font=footnotesize,labelfont=footnotesize}
  \includegraphics[width=\linewidth]{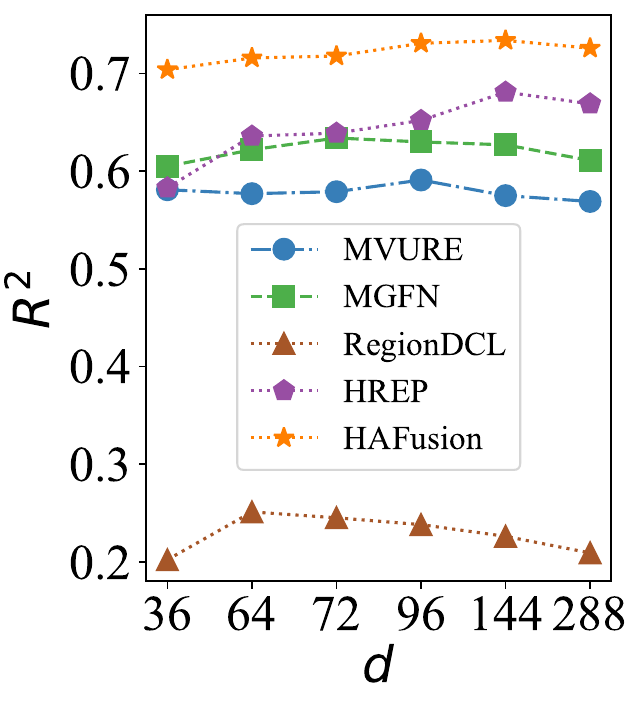}
  \vspace{-5mm}
  \caption*{(b) Crime}
\end{minipage}
\begin{minipage}[t]{0.32\columnwidth}
\captionsetup{font=footnotesize,labelfont=footnotesize}
  \includegraphics[width=\linewidth]{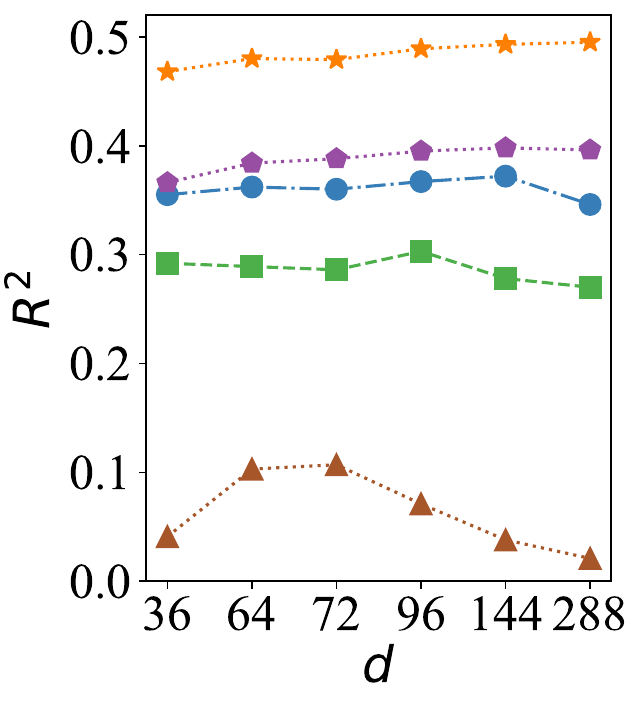}
  \vspace{-5mm}
  \caption*{(3) Service Call}
\end{minipage}
\caption{Impact of $d$ (NYC).}
\label{fig:ablation_dim}
\end{figure}

\textbf{Impact of the region embedding dimensionality ${d}$.}
 We study the impact of $d$ by using the same $d$ values across all models. We vary $d$ from 36 to 288 and use the learned region embeddings for three downstream prediction tasks like in the previous experiments. As Fig.~\ref{fig:ablation_dim} show, \model~consistently outperforms all competitors across all values of $d$ in terms of the prediction accuracy under the $R^2$ measure (results on the MAE and RMSE measures show similar patterns and are omitted for conciseness). The model prediction accuracy grows with $d$ initially. It starts to drop as $d$ continues to grow, where we conjecture overfitting may have occurred.

 The peak performance of different models is observed at different $d$ values. Our model \model~performs better across three tasks when $d$ is between 144 and 288. As a larger $d$ value also translates to longer running time, we default our model to using $d = 144$ to strike a balance between learning effectiveness and efficiency. The optimal choice of $d$ for all baseline models is consistent with the respective original papers. For example, the recommended value of $d$ for MGFN is 96, where the model yields the best performance on the check-in and service call prediction tasks and the second-best performance on the crime prediction task. We thus have used this $d$ value for MGFN by default.   

\begin{table}[htbp] \scriptsize
\caption{Impact of $\#layers$ (NYC)}
\vspace{-2mm}
\label{tab:para_layers}
\begin{center}
\setlength{\tabcolsep}{6pt}
\renewcommand{\arraystretch}{1} 
\resizebox{\columnwidth }{!}{
\begin{tabular}{l*{5}{c}}
    \hlineB{3}
    \#layers& 1 &2 & 3 & 4 & 5\\
    \hline & \\[-2ex]
    Prediction Task&$R^{2} \uparrow$ &$R^{2} \uparrow$ &$R^{2} \uparrow$ &$R^{2} \uparrow$ &$R^{2} \uparrow$ 
    \\ \hline & \\[-2ex]
    
   Check-in & 0.838  & 0.839 & \textbf{0.844} & 0.842 & 0.840 \\
    
    Crime & 0.721 & 0.725  & \textbf{0.734} & 0.729 & 0.706 \\

    Service Call & 0.483 & 0.486 & \textbf{0.493} & 0.492 & 0.481 \\
    \hline \\[-2ex]
\end{tabular}
}
\end{center}
\vspace{-1mm}
\end{table}

\textbf{Impact of the number of \moduleD~layers in \fusion\ $\#layers$.}   The number of layers in \fusion\ reflects the depth of our model. We vary it from 1 to 5. As Table.~\ref{tab:para_layers} shows, our model performance initially improves as $\#\textit{layers}$ increases but starts to decline when $\#\textit{layers}$ exceeds 3.
Adding more layers can help the model better capture intricate patterns, as the model can learn different levels of feature representations. However, when there are too many layers, the model is prone to overfitting due to its high capacity to memorize the training data, resulting in poor generalization to unseen data. We thus have set $\#\textit{layers}$ as 3 by default.
\section{Conclusion}
We proposed a model named~\model\ for urban region representation learning by leveraging human mobility, POI, and land use features. To learn region embeddings from each region feature, we proposed a hybrid attentive feature learning module named \learning\ that captures the abundant correlation information between different regions within a single view and across different views. To fuse region embeddings learned from different region features, we further proposed a dual-feature attentive fusion module named \fusion\ that encodes higher-order correlations between the regions. We perform three urban prediction tasks using the embeddings learned by \model. The results show that our learned embeddings lead to consistent and up to 31\% improvements in prediction accuracy, comparing with those generated by the state-of-the-art models. Meanwhile, our \fusion\ module helps improve the quality of the learned embeddings by up to 36\% when integrated into existing models.
\section{acknowledgment}
This work is partially supported by Australian Research Council (ARC) Discovery Projects DP230101534 and DP240101006.

\balance
\bibliographystyle{IEEEtran}
\bibliography{main}
\end{document}